\documentclass[10pt,twocolumn,letterpaper]{article}

\usepackage{wacv}
\usepackage{times}
\usepackage{epsfig}
\usepackage{graphicx}
\usepackage{amsmath}
\usepackage{amssymb}

\usepackage{multirow}
\usepackage{subcaption}
\usepackage{enumitem}
\usepackage{color}
\usepackage{cite}
\usepackage{makecell}
\usepackage{booktabs}

\definecolor{ao}{rgb}{0.0, 0.5, 0.0}
\usepackage{nopageno}

\wacvfinalcopy % *** Uncomment this line for the final submission

\ifwacvfinal
\def\assignedStartPage{9876} % *** Enter the assigned starting page number (instead of 9876)
\fi

\ifwacvfinal
\usepackage[breaklinks=true,bookmarks=false]{hyperref}
\else
\usepackage[pagebackref=true,breaklinks=true,colorlinks,bookmarks=false]{hyperref}
\fi

\hypersetup{
    colorlinks=true,
    linkcolor=red,
    urlcolor=magenta,
    citecolor=green,
}

\ifwacvfinal
\else
\fi

\begin{document}

%%%%%%%%% TITLE
\title{Revisiting Street-to-Aerial View Image Geo-localization \\
and Orientation Estimation}

\author{Sijie Zhu, Taojiannan Yang, Chen Chen\\
Department of Electrical and Computer Engineering, University of North Carolina at Charlotte \\
{\tt\small \{szhu3, tyang30, chen.chen\}@uncc.edu}
}

\maketitle
%\thispagestyle{empty}

%%%%%%%%% ABSTRACT

\begin{abstract}
Street-to-aerial image geo-localization, which matches a query street-view image to the GPS-tagged aerial images in a reference set, has attracted increasing attention recently. In this paper, we revisit this problem and point out the ignored issue about image alignment information. We show that the performance of a simple Siamese network is highly dependent on the alignment setting and the comparison of previous works can be unfair if they have different assumptions. Instead of focusing on the feature extraction under the alignment assumption, we show that improvements in metric learning techniques significantly boost the performance regardless of the alignment. 
%no matter the alignment is available or not. 
Without leveraging the alignment information, our pipeline outperforms previous works on both panorama and cropped datasets. Furthermore, we conduct visualization to help understand the learned model and the effect of alignment information using Grad-CAM. With our discovery on the approximate rotation-invariant activation maps, we propose a novel method to estimate the orientation/alignment between a pair of cross-view images with unknown alignment information. It achieves the state-of-the-art result on the CVUSA dataset.
%(\textbf{from 60.1\% to 70.4\% top-1 accuracy on CVUSA}).
%Moreover, we conduct visualization to help understand the learned model and the effect of alignment information. We find the activation map can be independent of the alignment of the training data and is able to provide additional information for orientation estimation. 
\end{abstract}

\section{Introduction}
\label{sec:introduction}
Image based geo-localization aims at providing image-level GPS location by matching a query street/ground image with the GPS-tagged images in a reference dataset. Instead of relying on street-view images \cite{zamir2014image} as the reference dataset, street-to-aerial geo-localization \cite{lin2013cross} leverages GPS-tagged aerial-view images as the reference, given their more complete coverage of the Earth than street-view images. In the early work \cite{lin2013cross}, street-to-aerial geo-localization is proposed to coarsely localize isolated images where no nearby ground-level image is available. With emerging deep learning techniques, recent works \cite{cvm,vo} are able to achieve high geo-localization accuracy on city-scale datasets such as CVUSA \cite{Zhai} and Vo \cite{vo}. In scenarios where GPS signal is noisy \cite{phone}, image geo-localization can provide additional information to achieve fine-grained localization. Street-to-aerial geo-localization is also proved effective on city-scale street navigation \cite{Li_2019_ICCV}. These practical applications make cross-view image geo-localization an important and attractive research problem in the computer vision community. 
%Given a street view image as a query, our goal is to determine its GPS location by matching it with the GPS-tagged aerial images in a reference dataset. In contrast to the traditional pixel-wise image geo-registration approach \cite{sheikh2003geodetic}, which requires accurate telemetry and sensor model, image geo-localization based on image matching is an image-level coarse geo-localization, and has attracted growing interest since it is free from the constraint of requiring the metadata. 
\iffalse
\begin{figure}[htbp]
\begin{center}
\includegraphics[width=0.99\linewidth]{fig_new/Figure1_overlook.pdf}
\vspace{-0.4cm}
\end{center}
\caption{An overview of the proposed framework for cross-view image geo-localization and orientation estimation.}
\vspace{-0.4cm}
\label{fig:1}
\end{figure}
\fi

\begin{figure}[t]
%\vspace{-0.4cm}
\begin{center}
%\vspace{-0.8cm}
\includegraphics[width=0.85\linewidth]{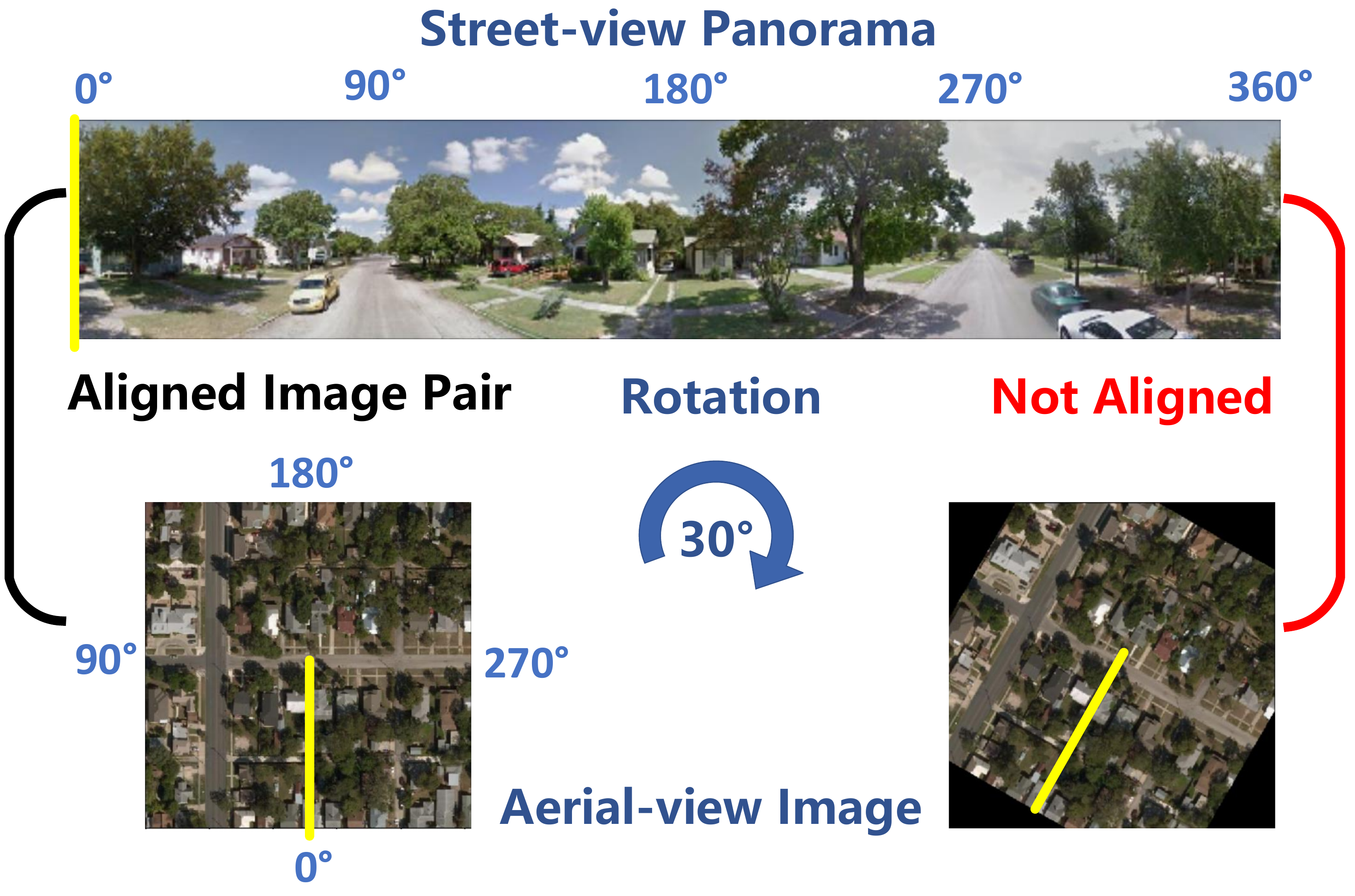}
\vspace{-0.5cm}
\end{center}%\vspace{-1em}
\caption{An example of alignment between street and aerial views. Yellow line denotes the South direction ($0^{\circ}$).}
%\vspace{-1.cm}
\label{fig:alignment}
\end{figure}

Recently, a number of works \cite{vo,cvm,lending,UCF,reweight} are proposed to address the street-to-aerial geo-localization problem and the performance seems to be improved significantly. A key ingredient of the prior work is to learn a feature embedding, such that the distance of a matched pair of images is small whereas the distance of the unmatched pair is large in this feature space, which is also known as metric learning. However, existing works have different settings about the alignment between street and aerial views (Fig. \ref{fig:alignment}), which can lead to unfair comparison. CVMNet \cite{cvm} is trained with randomly rotated aerial images without leveraging alignment information in the training set, thus is applicable for inference scenarios where alignment is not available, e.g. images on social media. \cite{vo,reweight} take advantage of the alignment information as additional supervision, but they do not assume the inference image pair to be well aligned. \cite{cvm,vo,reweight} also apply their algorithms to cropped street view image dataset, i.e. Vo dataset \cite{vo} (Fig. \ref{fig:2}), where panorama street view is not available. On the contrary, \cite{transport,UCF,NIPS2019_9199} aim at image or feature transformation from one view to the other, which is \textit{dependent on the geometric relationship between two views}. Although these techniques can boost the retrieval performance, but their presupposition does not hold when the alignment is not available for inference or only cropped street view images are provided. It can be unfair to claim state-of-the-art (SOTA) performance based on comparison with methods which do not have alignment assumption like \cite{cvm}. 
%A straightforward question is how the alignment information would affect the retrieval model in terms of performance (Table \ref{table:rotation}) and beyond (Fig. \ref{fig:rotation})? Without assuming that the inference image pairs are aligned, how to effectively improve the retrieval performance? Is it possible to estimate the alignment information when no explicit supervision is given?
These challenges motivate us to study three problems: \textit{1) how the alignment information would affect the retrieval model in terms of performance (Table \ref{table:rotation}) and beyond (Fig. \ref{fig:rotation}); 2) without assuming the inference image pairs are aligned, how to effectively improve the retrieval performance; 3) is it possible to estimate the alignment information when no explicit supervision is given?}

\begin{table}[htbp]
%\vspace{0.2cm}
\begin{center}
%\small
\begin{tabular}{c|cc}
%\toprule
\hline
\multirow{2}*{Validation} & \multicolumn{2}{c}{Training} \\
\cline{2-3}
~ & Aligned & {\textbf{Rotate}}\\
\hline
\hline
{Aligned} & 60.1\%& 43.7\%\\
\hline
{\textbf{Rotate}} & 13.5\%  & 44.2\%\\ 
\hline
%\bottomrule
\end{tabular}
\end{center}
\vspace{-0.5cm}
\caption{\small{Top-1 recall accuracy of Siamese VGG with different alignment settings (Aligned or Randomly Rotated (\textbf{Rotate} for short)) for training and validation sets on CVUSA dataset.}}
\label{table:rotation}
\end{table}

In this paper, we revisit the street-to-aerial view geo-localization problem and answer those three questions. We first shed light on the effect of alignment between two views, which is usually ignored in the discussion of previous works, by conducting ablation studies with a simple Siamese \cite{Siamese} network under different alignment settings. 
Without the alignment assumption in inference phase, improving metric learning is a promising direction, which is explored but not well-exploited in \cite{reweight}. By identifying the unique challenges of cross-view image matching in the context of metric learning, we further show that two techniques (global mining strategy and a new loss function) specifically tailored to the challenges can significantly improve the retrieval performance regardless of the alignment setting. Moreover, we leverage visual explanation \cite{gradcam} to investigate how the image matching model works for cross-view. \textit{Our observation reveals that the activation map of the matching model can provide geometric information which is independent of the alignment of the training data.} Inspired by this observation, we propose a novel orientation estimation method which significantly outperforms the existing approaches.
The main contributions of this paper are three-fold:
\begin{itemize}[leftmargin=*]
    \item We provide an in-depth analysis on image alignment, which is ignored by previous works, for cross-view matching. Ablation study and visual explanation lead to a key observation -- the alignment has a great impact on the retrieval performance. It provides valuable information for designing robust and general frameworks, and presenting fair comparisons with prior work.
    %This could serve as a good practice 
    
    %Ablation study and visual explanation are conducted to illustrate the effect of alignment information.
    
    %We point out the issue of alignment which is ignored by previous works. Comparison without clear claim about the alignment setting can be unfair. Ablation study and visual explanation are conducted to illustrate the effect of alignment information on cross-view matching.
    %\vspace{-0.2cm}
    \item We show that improvements on metric learning techniques constantly boost the retrieval performance regardless of the alignment information. Our specifically designed pipeline achieves the state-of-the-art results on two benchmarks when no assumption is made on the alignment of the inference set.
    %\vspace{-0.2cm}
    \item We discover that the orientation information between cross-view images can be estimated when the alignment is unknown. The proposed orientation estimation method outperforms previous methods without explicit supervision.% during training.
\end{itemize}
%Favorable experiment results on orientation estimation support our argument.
%-------------------------------------------------------------------------
\section{Related Work and Motivation}
\label{sec:related}
\subsection{Geo-localization}
\label{sec:geolocalization}
Recent works for cross-view geo-localization \cite{vo,cvm,lending,UCF,reweight} are all based on Siamese networks \cite{Siamese}, while they build their pipelines on different baselines with different settings. Vo \textit{et al.} \cite{vo} first propose a Siamese network with exhausted triplet loss \cite{facenet} for city-scale geo-localization with cropped street-view images and satellite images. Their results suggest that an auxiliary orientation regression task can further improve the performance. Hu \textit{et al.} \cite{cvm} propose CVMNet which combines Siamese VGG \cite{vgg} and NetVLAD \cite{netvlad} along with a modified version of the triplet loss. Liu \textit{et al.} \cite{lending} leverage the alignment information between street and aerial views and improve the performance by adding orientation information in the input. Krishna \textit{et al.} \cite{UCF} utilize GANs to generate image from one view to the other and adopt feature fusion to achieve higher accuracy. Shi \textit{et al.} \cite{transport} aim to find the optimal feature transformation between two views based on the geometric prior knowledge. Cai \textit{et al.} \cite{reweight} put more weight on the hard samples in an online negative mining manner and use a stronger backbone (ResNet \cite{resnet} with attention) to achieve better result. \textit{Although progress has been made for cross-view geo-localization, existing works %\cite{UCF,transport} 
with different settings on the cross-view alignment (discussed in Section 2.2) can lead to unfair comparison.} 
%Besides, the power of superior metric learning techniques is also not well explored.

\subsection{Alignment Setting}
CVMNet \cite{cvm} is trained on randomly rotated aerial images, in which case the alignment information is not available, thus resulting in a general framework for cross-view image matching. \cite{vo,reweight} leverage the alignment information in training set by adding a regression task for orientation prediction as additional supervision, while they are applicable for unaligned inference image pairs. However, \cite{lending} takes advantage of the alignment information by adding orientation as an auxiliary input, so the orientation is also required as input for inference images. The image/feature transformation in \cite{UCF,transport} also relies on the accurate geometric relationship between two views. These methods may not work well when the alignment between street and aerial views is not available or only cropped street view images are provided, which is often the case in real-world applications. It is clear that more supervision can result in better performance, but existing works \cite{UCF,lending} fail to make the effect of alignment information very clear in their comparisons. As shown in Table \ref{table:rotation}, the alignment information actually has a great impact on the accuracy.

Fig. \ref{fig:alignment} shows a graphical illustration of the alignment information of cross-view images. To investigate its impact on geo-localization performance, in Table \ref{table:rotation} we report the top-1 recall accuracy of Siamese-VGG with different alignment settings for training and validation sets on CVUSA. Training with randomly rotated aerial images (the alignment information is therefore not available) yields a performance drop on the aligned validation set (from 60.1\% to 43.7\%) compared with training with aligned images.
But this trained model is able to perform well on the randomly rotated validation set (43.7\% vs 44.2\%).
On the other hand, the model trained with aligned images has an extremely low top-$1$ accuracy ($13.5\%$) on the randomly rotated validation set. These results indicate that training without alignment information makes the model generalize better. Effect beyond performance will be discussed in Section \ref{sec:alignment}.

\subsection{Metric Learning}
\label{sec:metric}
Unlike feature transformation, metric learning techniques are \textit{independent of the alignment assumption}. General metric learning aims to learn an embedding space where positive samples (matched pairs) are close to each other, while negative samples (unmatched pairs) have a large distance between each other. Recent works~\cite{lending, UCF} usually adopt the loss from~\cite{cvm}, i.e. a modified triplet loss~\cite{facenet}, along with the within batch negative mining \cite{facenet} or assigning more weight on hard negative samples in a mini-batch \cite{reweight}. Although the common techniques for metric learning, e.g. triplet loss and hard negative mining, are employed in recent geo-localization methods, the unique challenges of cross-view geo-localization are not specifically addressed. %\cc{We say the challenges of geo-localization in Contribution (2), so I think we should say ``the unique challenge" instead of ``characteristic" or ``senario" here? Also, I add a paragraph title in the next paragraph to indicate the two challenges.} 
%In the following, we elaborate on these challenges. %\par

\textbf{Challenges for cross-view geo-localization.} For street-to-aerial view matching, most of the time, there is only one matched (\textbf{positive}) aerial-view image for the query street-view image from the same location. On the contrary, all the aerial images from other locations are considered as \textbf{negative} samples. As a result, there is a significant imbalance between positive and negative pairs. Therefore, different from the Facenet~\cite{facenet} dataset which contains about 20 different images for one face ID, the number of positive samples for an anchor street-view image is very limited in geo-localization, i.e. only one.
%\footnote{We argue even though one can do data augmentation, e.g. rotation and cropping on the aerial image at one location, the semantic information would not increase much.}. 
The boundary between positive pairs at different locations is difficult to estimate by only one sample in the embedding space. Since most existing methods follow the form of triplet loss which gives the same weight for positive and negative samples, \textit{the large imbalance between positive and negative samples inspires us to design a better loss function in \textbf{Section~\ref{sec:loss}} for this task.}

As the training accuracy increases, most training samples are correctly handled and have little contribution to the overall loss \cite{facenet}, therefore hard negative mining~\cite{facenet} is necessary. Several geo-localization methods~\cite{cvm, lending, UCF, reweight} use a small batch size to fit the high resolution images in memory. Although online negative mining within mini-batch is employed, it does not work well when the training accuracy is high since almost no hard pairs can be found in a mini-batch. \textit{We solve this problem in \textbf{Section~\ref{sec:mining}} by introducing a global mining strategy}.% with little additional computational cost.}

\subsection{Orientation Estimation}
Several cross-view orientation estimation methods have been proposed along with geo-localization. Vo \textit{et al.} \cite{vo} predict the rotated angle of aerial image by adding an auxiliary \textit{supervised} regression task, with the goal of improving localization accuracy. The auxiliary regression sub-network is able to coarsely predict the orientation angle between street and aerial view images. Zhai \textit{et al.} \cite{Zhai} first predict the semantic segmentation map of street-view panorama from aerial image by learning a transformation matrix between two views. Then the segmentation map is matched with the one generated directly from street-view crop in a sliding-window manner to find the best matching angle. To learn the cross-view transformation matrix, their model composing of three sub-networks has to be trained on \textit{well-aligned} street and aerial image pairs. A pre-trained segmentation model is also required as additional supervision. In summary, these methods~\cite{vo, Zhai} require \textit{explicit supervision (i.e. image alignment)} on the training data to train their models for orientation estimation/prediction. \textit{We propose an orientation estimation approach (\textbf{Section~\ref{sec:orientation}}) does not rely on the alignment information for training, yet is able to achieve superior performance.}
%------------------------------------------------------------------------

\section{Retrieval Framework}
\subsection{Baseline Architecture}

We adopt a simple Siamese-VGG as our baseline architecture with the loss function of \cite{cvm}. Given a set of training pairs including street-view images $x_{i}$ and their corresponding aerial images $y_{i}$, our framework learns two mapping functions, $f(X,\Theta_{x})$ and $g(Y,\Theta_{y})$, which map each input pair $(x_{i},y_{i})$ into a $K$-dimensional space. The goal is to find the best $\Theta_{x}$ and $\Theta_{y}$ so that $g(y_{i},\Theta_{y})$ is the nearest neighbor of $f(x_{i},\Theta_{x})$ in the embedding space. %As shown in Fig.~\ref{fig:2}, we use a simple Siamese framework with two backbone CNNs which respectively corresponds to the function $f$ and $g$.
As shown in Fig.~\ref{fig:2}, we use the same backbone architecture (e.g. VGG) for both views, i.e. $f=g$. Due to the significant visual difference between two views, the convolutional layers denoted by two separate CNNs in Fig.~\ref{fig:2} are trained without sharing weights in order to extract different low-level features from two views. On the other hand, fully connected layers of the two streams share weights, since the high-level feature (e.g. semantic information) is similar in both views. $L_{2}$ normalization is used for the output feature vectors of both views.

\begin{figure}[h]
\begin{center}
%\vspace{-0.6cm}
\includegraphics[width=0.98\linewidth]{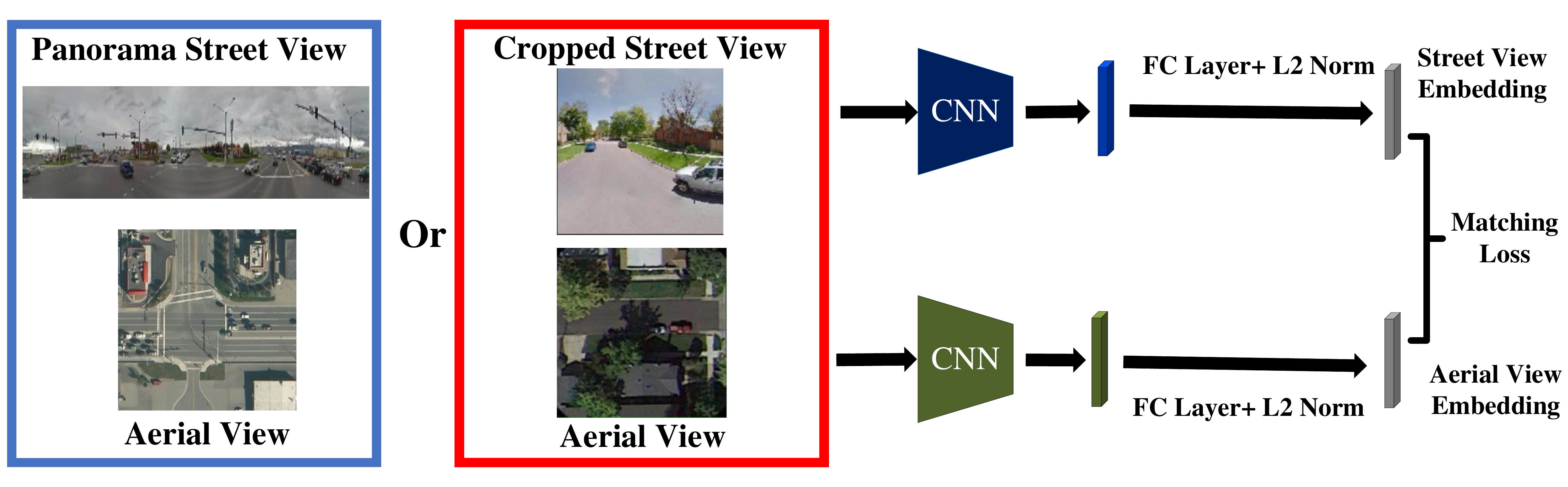}
%note: remove GAP
\vspace{-0.3cm}
\end{center}
\caption{The overall framework of our method.}
\vspace{-0.3cm}
\label{fig:2}
\end{figure}

\subsection{Binomial Loss}
\label{sec:loss}
%\paragraph{Triplet Loss.} 
Triplet loss has been widely used in a variety of image matching tasks, including face clustering \cite{facenet}, person re-identification (Re-ID) \cite{hermans2017defense,liao2017triplet} and image retrieval \cite{yu2018correcting,gordo2017end}. The idea is to teach the network to pull positive samples close to the anchor, while at the same time, push negative samples far away by a margin $m$. The triplet loss function is written as:
\begin{equation}
\small
    \label{eq:1}
    L = \frac{1}{N}\sum^{N}_{i}max(0,d^{p}_{i}-d^{n}_{i}+m).
\end{equation}
For $N$ triplets in a mini-batch, $d^{p}_{i}$ and $d^{n}_{i}$ denote the distance between the $i$-th anchor and its corresponding positive and negative samples. Squared and non-squared Euclidean distances are commonly used in triplet loss. Different from the hard-margin function in Eq. (\ref{eq:1}), Vo \textit{et al.} \cite{vo} propose to use a soft-margin function $\sigma(d) = log(1+exp(d))$ in the triplet network for geo-localization, where $d = d^{p}_{i}-d^{n}_{i}$. The soft-margin loss has a Sigmoid gradient function which is more smooth than hard margin. Moreover, Hu \textit{et al.} \cite{cvm} add a parameter $\alpha$ to form a weighted soft-margin loss:
\begin{equation}
\small
    \label{eq:2}
    L = \frac{1}{N}\sum^{N}_{i}\sigma(\alpha(d^{p}_{i}-d^{n}_{i})), \quad \alpha>0.
\end{equation}
The parameter $\alpha$ should be tuned to find a suitable balance for the gradients of easy and hard samples. However, $d^{p}_{i}$ and $d^{n}_{i}$ always have the same weight $\alpha$ in Eq.~\ref{eq:2}, resulting in the same magnitude of gradient. That means positive and negative samples will be pulled and pushed in the same manner. For our positive and negative imbalanced case, adjusting different weights for positive and negative samples can lead to a loss function which better alleviates this problem.
%\paragraph{Different Weights.} 
Yi \textit{et al.} \cite{yi2014deep} have successfully utilized binomial deviance loss for person Re-ID. It is formulated as:
\begin{equation}
\small
    \label{eq:3}
    L = \frac{1}{N_{p}}\sum^{N_{p}}_{i}\sigma(-\alpha (s^{p}_{i}-m))+ \frac{1}{N_{n}}\sum^{N_{n}}_{i}\sigma(\alpha(s^{n}_{i}-m)).
\end{equation}
Here $s^{p}_i$ and $s^{n}_i$ denote the \textbf{cosine similarity} between the $i$-th anchor and its positive and negative samples. $N_p$ and $N_n$ represent the number of positive and negative pairs, respectively. Although the original loss function uses the same parameters $\alpha$, $m$ for positive and negative samples, we make use of this formula while assign different $\alpha_{p}$, $\alpha_{n}$ and $m_{p}$, $m_{n}$. We use the normalized $\sigma$ formulated as $\hat{\sigma}(\alpha,d) = \frac{1}{\alpha}\sigma(\alpha d)$. Then the gradient is given by $ \partial\hat{\sigma}(\alpha,d)/\partial d = 1/(1+exp(-\alpha d))$, and our loss function becomes:
\begin{equation}
\small
    \label{eq:4}
   L = \frac{\sum^{N_{p}}_{i}\sigma(-\alpha_{p} (s^{p}_{i}-m_{p}))}{\alpha_{p} N_{p}}+ 
   \frac{\sum^{N_{n}}_{i}\sigma(\alpha_{n}(s^{n}_{i}-m_{n}))}{\alpha_{n} N_{n}}.
\end{equation}
By setting a large $\alpha_{n}$, the gradients of negative pairs drop fast as $s^{n}$ decreases from $m_{n}$, which means only to push negative samples away by a small distance. However, with a small $\alpha_{p}$, the gradients of positive pairs drop slowly as $s^{p}$ increases from $m_{p}$, resulting in pulling positive samples to the anchor until $s^{p}_i$ is much greater than $m_{p}$. When positive samples are much fewer than negative samples, as in cross-view geo-localization with only one positive match, \textbf{it would be easier to pulling the only matched sample close to the anchor rather than pushing all negative samples away}. Therefore, we assign a much smaller value to $\alpha_{p}$ than $\alpha_{n}$ to validate this idea. To avoid too many hyper-parameters, we simply set $m_{p}$ and $m_{n}$ as the average values of $s^{n}$ and $s^{p}$.

\begin{figure*}[htbp]
\begin{center}
%\vspace{-0.5cm}
\includegraphics[width=0.9\linewidth]{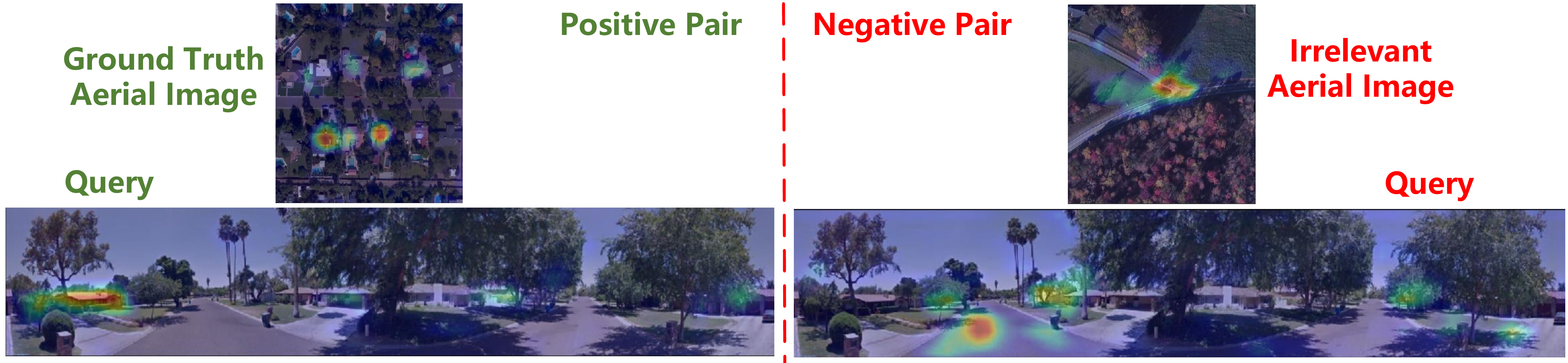}
\vspace{-0.5cm}
\end{center}
\caption{Grad-CAM activation map (overlaid on images) of our baseline on positive (left) and negative (right) image pairs.}
\label{fig:gradcam}
\end{figure*}
\begin{figure*}[htbp]
\begin{center}
%\vspace{-0.3cm}
\includegraphics[width=0.9\linewidth]{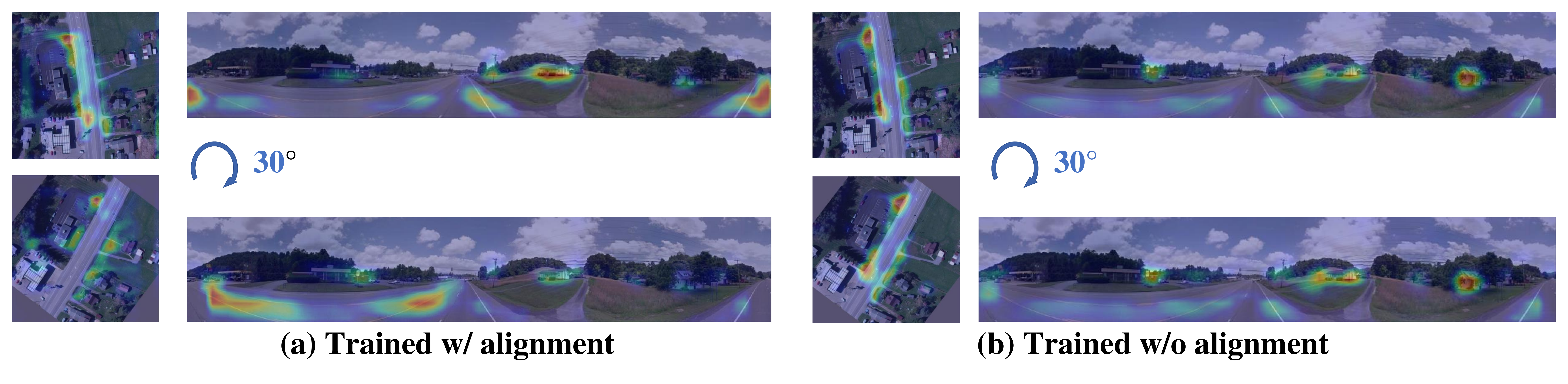}
\vspace{-0.65cm}
\end{center}
\caption{Grad-CAM activation maps (overlaid on images) of our Siamese VGG baseline trained on (a) aligned and (b) randomly rotated aerial images.}
\label{fig:rotation}
\end{figure*}

\subsection{Global Mining Strategy}
\label{sec:mining}

As the training accuracy increases, most of the negative samples contribute zero loss and the convergence becomes slow. In \cite{cvm}, mining the hardest triplet in a batch helps when there exists some hard negative samples, but it does not apply when a small batch size, e.g. 12 pairs (24 images), is used due to the high resolution images, e.g. panoramic street-view images. To find the exact hardest negative sample in a subset with $N_{m}$ pairs, the embedding vectors of samples in the subset have to be updated every $p$ steps ($p$ can be adjusted to find the best trade-off between computation and accuracy). To avoid large computational cost of this offline update strategy, our online mining computes the output vectors of each batch in the back-propagation step and saves them into a mining pool in an FIFO (first in first out) manner. Although the vectors in the mining pool have the values of several steps before, the delay is always less than one epoch. This strategy generates approximate hard negative samples in each step for every sample in a batch with negligible additional computation. When the training set is small, we save the vectors of the whole training set such that the global hard negative samples can be found. For each positive pair, we randomly select one of the $r$ hardest negative samples from the mining pool.

\section{Alignment Analysis}
\subsection{Visual Explanation}
\label{sec:alignment}
It is evident that using the cross-view alignment information can improve the performance, but how the alignment information may affect the geo-localization model?
In addition to the performance (Table \ref{table:rotation}), we take a closer look at the trained model with or without alignment, through the lens of Grad-CAM \cite{gradcam} -- a widely used visualization technique, in order to understand how the model actually works. The class activation map generated by Grad-CAM highlights the important regions which contribute the most to the final similarity between two images. Specifically, the class activation map is obtained by computing the gradient till the last convolution layer from the inner product of two views' embedding features without $L_{2}$ normalization \cite{zhu2019visual}.

As shown in Fig. \ref{fig:gradcam}, the activation map of the same query image can be dramatically different when the retrieved image is different. For the positive image pair in this example, the model mainly focuses on the discriminative objects in both views, i.e. \textit{houses}. However, for the negative pair, the model highlights the \textit{trees and roads} areas in the query image, as there is no house in the retrieved image. The visual explanation further demonstrates that the similarity score predicted by the metric learning model is mainly based on similar patterns in different views, and the activated regions of two views are highly relevant to each other. A reasonable hypothesis is that the activated regions of two views are likely to be the same objects. \textit{However, what if the image pair is not aligned}? In Fig. \ref{fig:rotation}, we show the changes of activation map corresponding to a rotation of $30^{\circ}$ in the aerial image, using two different models: (a) baseline trained with aligned images, (b) baseline trained with unaligned images. For model (a), the activation map changes dramatically on both views when the aerial image is rotated (Fig. \ref{fig:rotation} (a)), because the model trained with aligned images relies on specific geometric relationship between two views. However, model (b) which is trained with randomly rotated aerial images only leverages similar patterns for matching (Fig. \ref{fig:rotation} (b)), thus leading to approximate \textit{rotation-invariant} activation map, which may provide geometric information for orientation (or camera pose) estimation (Section \ref{sec:orientation}).

\subsection{Orientation Estimation Approach}
\label{sec:orientation}
%\textbf{All these methods require the alignment information of the training data}. However, with the highly relevant regions in the approximate rotation-invariant activation map (Fig. \ref{fig:rotation} (b)), we are able to estimate the orientation information \textbf{without explicit supervision}. 
\begin{figure}[htbp]
%\vspace{-0.8cm}
\begin{center}
\includegraphics[width=0.98\linewidth]{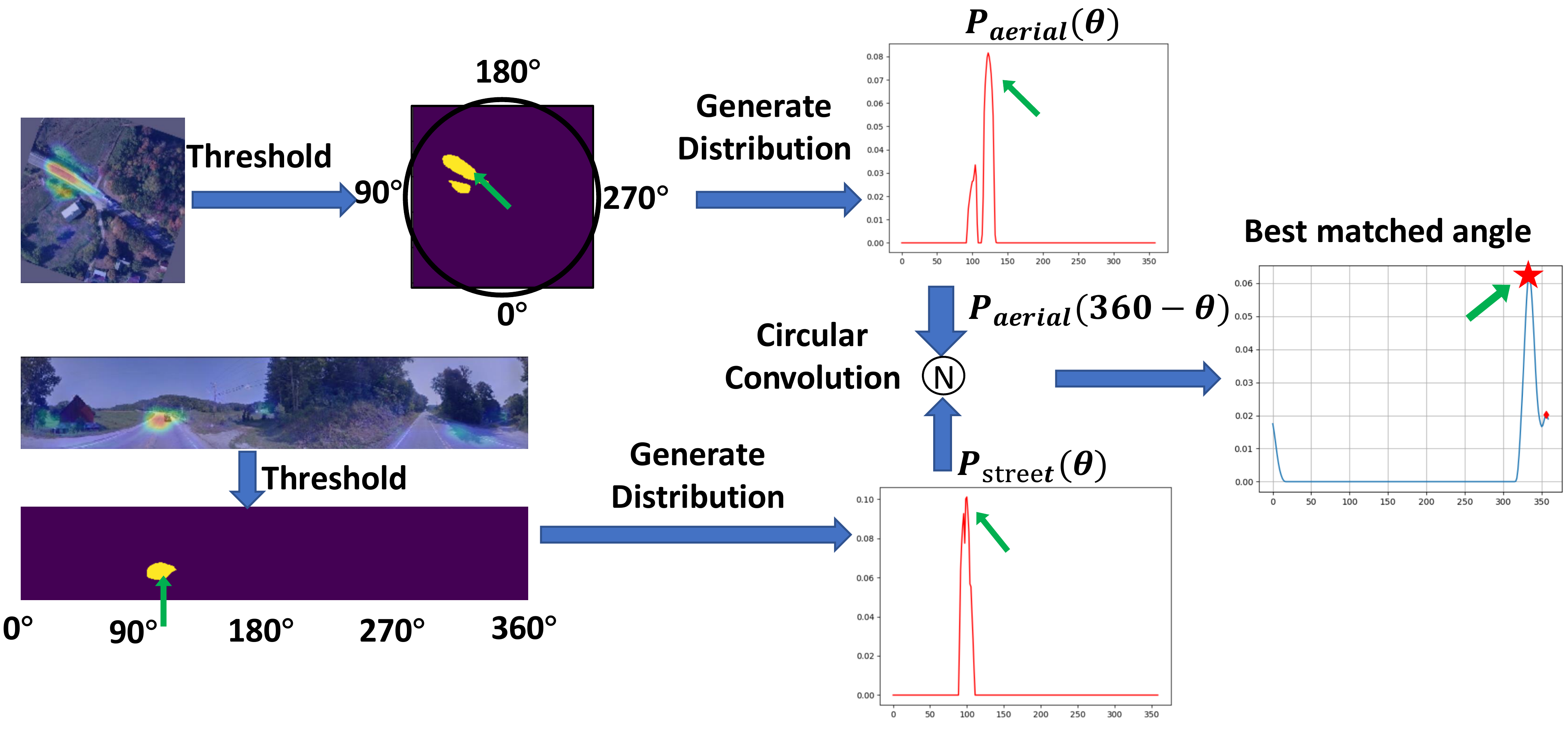}
\vspace{-0.3cm}
\end{center}
\caption{Framework of our orientation estimation approach. (Zoom in to view) %(Best viewed with zoom in on screen)
}
\label{fig:5}
\end{figure}
Since the most activated regions are likely to be the same objects, the angle distributions (Fig. \ref{fig:5}) of activated pixels from two views would be similar if the image pair is well aligned. A rotation of aerial image will cause the angle distribution to shift by a certain degree. With the observation of the \textit{approximate rotation-invariant activation maps} (Section~\ref{sec:alignment}), which are obtained from our matching
model trained with rotated aerial images, we propose to estimate the orientation of an unaligned image pair by matching the angle distributions of activated pixels in two views. As shown in Fig.~\ref{fig:5}, we first compute the activation maps using Grad-CAM and select pixels with values higher than a threshold. Then we calculate the angle distribution of the selected pixels in each view as $p_{street}(\theta)$ and $p_{aerial}(\theta)$. To find the angle $\phi$ so that $p_{aerial}(\theta+\phi)$ best matches $p_{street}(\theta)$, circular convolution of $p_{street}(\theta)$ and $p_{aerial}(360-\theta)$ is computed efficiently using the Fast Fourier Transform (FFT) algorithm as follows (\textcircled{\footnotesize{N}} denotes circular convolution):
\begin{equation}
\begin{aligned}
\small
    \label{eq:5}
    &p(\theta)= p_{street}(\theta) \quad \textcircled{\footnotesize N} \quad p_{aerial}(360-\theta) \\ = &\mathtt{ifft}(\mathtt{fft}(p_{street}(\theta))\mathtt{fft}(p_{aerial}(360-\theta))),
\end{aligned}
\end{equation}
where $\mathtt{fft}$ and $\mathtt{ifft}$ are the Discrete Fourier Transform (DFT) operation and inverse DFT operation.
Finally, the highest peak is selected as the predicted angle. \textit{Since the training image pairs are not required to be aligned, the proposed method does not leverage explicit supervision of orientation, which is a clear advantage as compared with previous supervised counterparts \cite{vo, Zhai}.}
%and the percentage of this peak value with respect to the sum of all peaks is considered as the prediction confidence. 

%-------------------------------------------------------------------------
\section{Experiment}
%\subsection{Dataset}
\textbf{Dataset.} We conduct experiments on two popular benchmark datasets, i.e. CVUSA \cite{Zhai} (panorama street view) and Vo \cite{vo} (cropped street view), see Fig.~\ref{fig:2}. These two datasets are city-scale benchmarks for cross-view geo-localization. The original CVUSA (Cross-View USA) \cite{workman2015wide} dataset contains more than $1$ million of ground-level and aerial images from across the US. Furthermore, Zhai \textit{et al.} \cite{Zhai} make use of the camera's extrinsic parameters to generate aligned pairs by warping the panoramas, resulting in 35,532 image pairs for training and 8,884 image pairs for testing. We use the same selected CVUSA \cite{Zhai} in our experiment.% and some examples are presented in Section \ref{sec:case}.

Vo \textit{et al.} \cite{vo} introduced a cross-view dataset which consists of about one million image pairs from 11 cities in the US. They randomly queried street-view panorama images and made several crops from each panorama with overhead images from Google Map. Small crops and overlapping images make this dataset more challenging. For a fair comparison, we follow the splitting in \cite{vo} using 8 cities for training and Denver city as the test set.

%\subsection{Implementation Details}
\textbf{Implementation Details.} For the proposed baseline, we adopt VGG-16 \cite{vgg} as the backbone architecture for both views with the loss function in Eq. \ref{eq:2} \cite{cvm} ($\alpha = 20$). Adam \cite{adam} optimizer with learning rate decay is used for training and the dimension of the embedding space is the same as CVMNet \cite{cvm}. %Random central crop is used for data augmentation, and 
Random aerial view rotation is adopted for training without alignment. The proposed mining strategy and simple backbone facilitate the training on a single GPU (Nvidia 1080Ti) with a small batch size, i.e. 12 pairs for CVUSA and 32 pairs for Vo. All experiments are implemented based on Tensorflow \cite{abadi2016tensorflow}. The mining pool is the whole training set for CVUSA and we use a subset of $10,000$ pairs for Vo with $r=100$. The mining pool is updated every epoch and the binomial loss is adopted after the distribution is stable (30 epochs). We set $\alpha_{p}=5$, $\alpha_{n}=20$ for both CVUSA and Vo. We simply set $m_{p}$, $m_{n}$ as $0$, $0.7$ based on the average values of the distributions of $s^{n}$ and $s^{p}$ as shown in Fig. \ref{fig:distribution}.

\subsection{Retrieval Performance}%{Comparison with Published Results}
\textbf{Evaluation Metrics.} For geo-localization, we use the top-$n$ recall accuracy as the evaluation metric on both datasets. For each query image, the retrieval is considered successful if the ground-truth reference is ranked within the top $n$ retrieved images. Instead of only providing the top-$1\%$ accuracy, we also report the top-1 accuracy to better demonstrate the performance of our approach. The top-$1\%$ accuracy is useful when the performance is poor, but it is not discriminative anymore if the top-$1\%$ is already higher than $95\%$ as in this paper. Since a high top-$1$ accuracy is the ultimate goal of geo-localization for practical applications, we highly recommend including the top-$1$ accuracy in future research.

We compare our method with existing methods on two datasets in Table \ref{table:accuracy}. We use the published results of \cite{workman2015wide,vo,cvm,UCF,lending,reweight} for top-$1\%$ accuracy comparison. The top-1 accuracy of some competing methods is not available. For CVM-Net \cite{cvm}, we reproduce their test result on CVUSA with their pre-trained model for top-1 and top-1\% accuracy\footnote{The top-1\% accuracy we generated using their model and code is $2\%$ higher than their published result in \cite{cvm}.}, while the result of CVM-Net on Vo is from their paper since the pre-trained model for this dataset is not provided.
%The top-1 to top-20 accuracy of our approach is presented in Section \ref{sec:ablation}.
\begin{table}[htbp]
\small
\begin{center}
\begin{tabular}{l|c|c|c|c}
\hline
\multirow{2}*{Method} & \multicolumn{2}{c|}{CVUSA} & \multicolumn{2}{c}{Vo}\\
\cline{2-5}
~ & Top-1\% & Top-1 & Top-1\% & Top-1 \\
\hline
Scott \cite{workman2015wide}\tiny{(ICCV'15)}& 34.3\%& -& 15.4\%  & -\\
Zhai \cite{Zhai}\tiny{(CVPR'17)}& 43.2\%  & - & - & -\\
Vo \cite{vo}\tiny{(ECCV'16)}& 63.7\%  & - & 59.9\%  & -\\
CVMNet \cite{cvm}\tiny{(CVPR'18)}& 93.6\% & 22.5\% & 67.9\% & -\\
Lending \cite{lending}\tiny{(CVPR'19)}& 93.19\% & 31.71\% & - & - \\
Reweight \cite{reweight}\tiny{(ICCV'19)}& 98.3\% & 46.0\% & 78.3\% & -\\
GAN \cite{UCF}\tiny{(ICCV'19)}& 95.98\% & 48.75\% & - & -\\
\hline
%Our baseline & \textbf{98.8\%} & \textbf{60.1\%} &	\textbf{84.3\%} & \textbf{7.9\%} \\
%\hline
%Our overall & \textbf{99.1\%} & \textbf{70.4\%} & \textbf{88.3\%} &\textbf{11.8\%}\\
\textbf{Ours} & \textbf{97.7\%} & \textbf{54.5\%} & \textbf{88.3\%} &\textbf{11.8\%}\\
\hline
\end{tabular}
\end{center}
\vspace{-0.4cm}
\caption{Top-1 and top-1\% recall accuracy comparison on CVUSA and Vo datasets.}
\label{table:accuracy}
%\vspace{-0.8cm}
\end{table}

\textbf{CVUSA.} 
As shown in Table \ref{table:accuracy}, the overall proposed pipeline significantly surpasses existing methods without the alignment assumption on inference set. Note that we do not adopt any recent strong network backbones as in \cite{reweight} (ResNet+attention), we believe our approach would gain extra improvement from using stronger network backbones. Our method also enjoys simple architecture and therefore is easy to reproduce and requires less computation compared with CVMNet. Moreover, our global mining strategy brings superiority on convergence speed for the training process (details in Section \ref{sec:ablation} and Fig. \ref{fig:convergence}). Please also refer to the \textbf{Appendix} for qualitative results and top-$n$ accuracy distribution.

%\vspace{0.2cm}
\textbf{Vo.} Vo \cite{vo} is a much more challenging dataset compared with CVUSA, as it contains around 10 times of images for validation and the cropped street-view images provide less information than panorama. Furthermore, the image/feature transformation based on geometric prior knowledge is infeasible for this case. Therefore, as shown in Table \ref{table:accuracy}, every method yields a much lower accuracy on Vo than that on CVUSA. Also, none of the existing methods reports top-$1$ accuracy. Again, the proposed framework outperforms all existing methods by a large margin (\textbf{$\bf{10\%}$ on top-$\bf{1\%}$ accuracy}).  
Note that \cite{vo,reweight} utilize the alignment information of the training data by adding a rotation angle classification task (an auxiliary task) to their matching frameworks to boost the geo-localization performance, but we do not use this technique in our method.

\iffalse
\paragraph{Case Study.}To further illustrate our matching result, we present four examples on CVUSA with positive sample ranking at 1, 5, 99 and 957 in Fig.~\ref{fig:case}. The similarity score (``probability'' in Fig. \ref{fig:case}) is also provided for all retrieved images. The successful case ranked at 1 contains many objects, e.g. \textit{houses}, with distinguishable features. However, the second one only contains \textit{road and trees} which are similar in all images. All the top 4 samples can be regarded as hard negative for this anchor. Extremely poor cases in the third and fourth row also contain no distinguishable objects. Other possible explanation is the dramatic appearance difference between two views (may be acquired at different seasons or illumination intensity).
\label{sec:case}
\begin{figure*}[htbp]
\begin{center}
\includegraphics[width=0.99\linewidth]{fig_new/Figure5_case.pdf}
\vspace{-0.4cm}
\end{center}
\caption{Image retrieval examples on CVUSA. For each query street view image, we show the top-5 retrieved aerial images. The ground truth is marked in green box. The red box marks the same aerial image as the ground truth. The last two rows show two examples where the ground truth images are not in the top-5 (i.e. rank=99 and rank=957) of the retrieved results.}
\label{fig:case}
\end{figure*}
\fi

\subsection{Orientation Estimation Performance}
\begin{figure}[hbtp]
\vspace{-0.2cm}
\begin{center}
\begin{subfigure}{0.5\textwidth}
\centering
\includegraphics[width=0.73\linewidth]{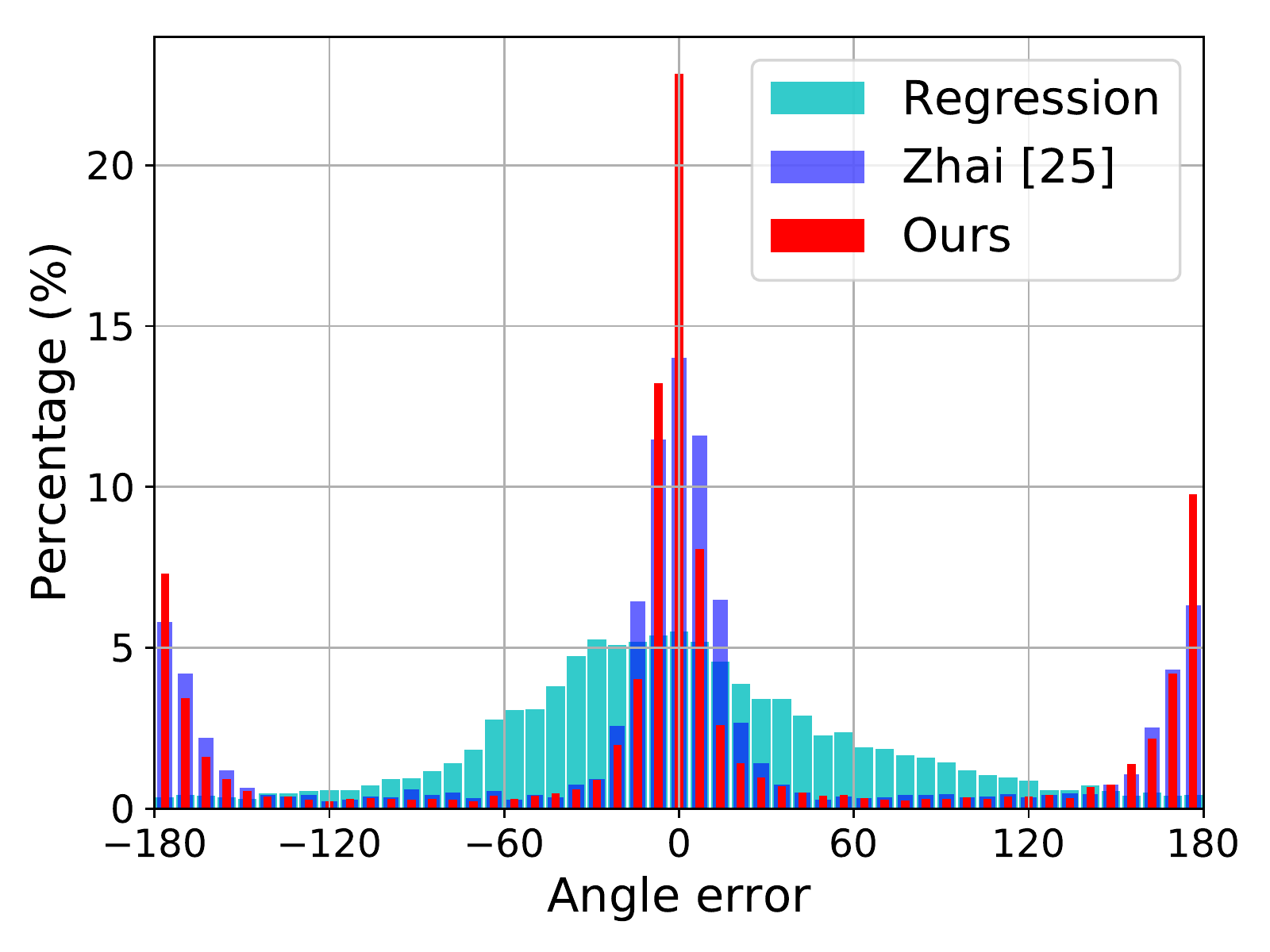}
\vspace{-0.2cm}
\caption{Our method vs. previous methods.}
\end{subfigure}
\begin{subfigure}{0.5\textwidth}
\centering
\includegraphics[width=0.73\linewidth]{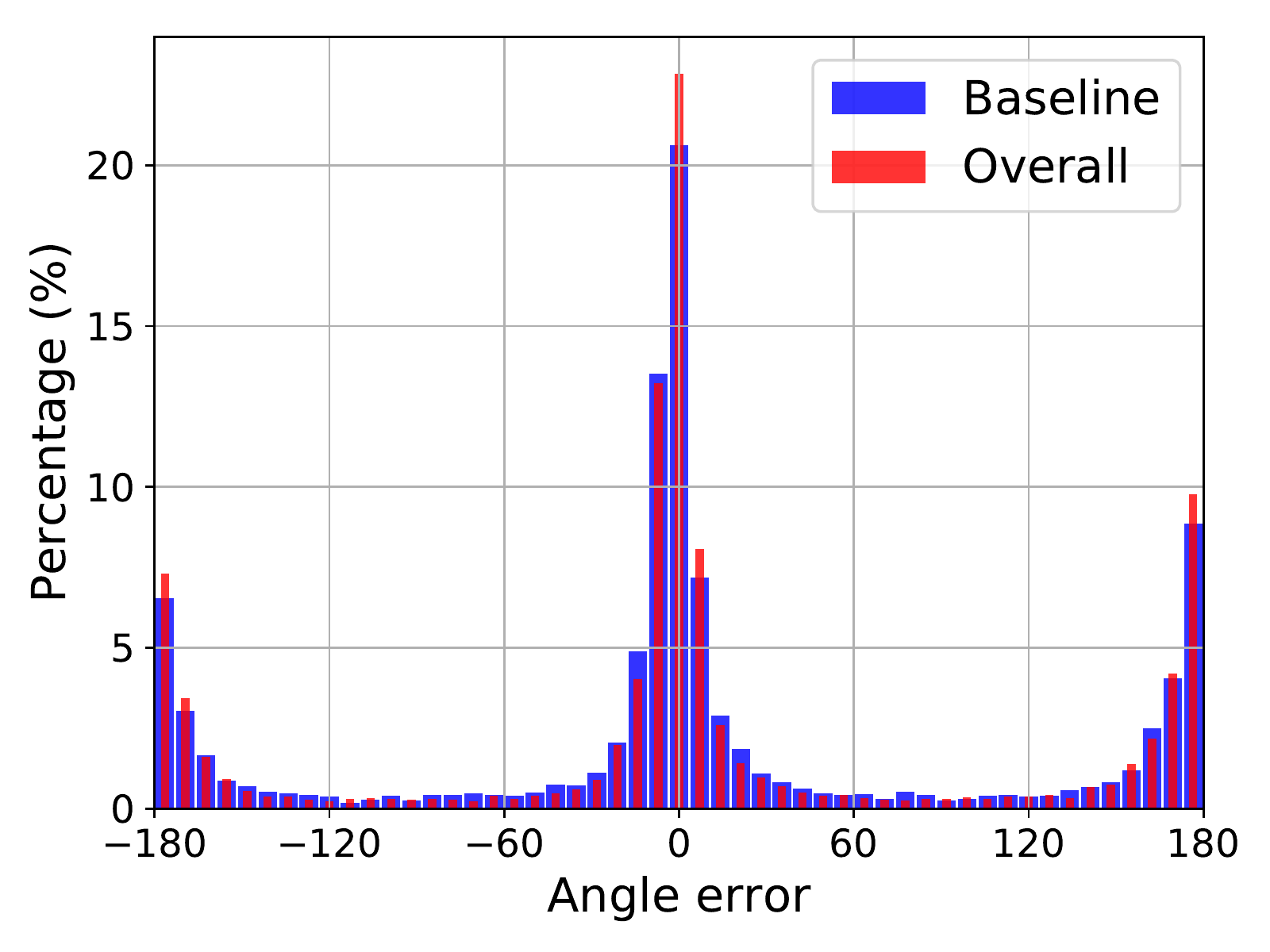}
\vspace{-0.2cm}
\caption{Overall model vs. baseline model.}
\end{subfigure}
\vspace{-0.2cm}
\caption{Error distribution of orientation estimation on CVUSA in percentage (best viewed in color with zoom in).}
\vspace{-0.7cm}
\label{fig:OE}
\end{center}
\end{figure}

As described in Section~\ref{sec:orientation}, our trained matching model is able to roughly predict the angle between paired aerial and street view images without any alignment information in the training set. We conduct experiments on CVUSA with randomly rotated aerial images and evaluate the distribution of angle prediction errors. For comparison, we train our baseline with an auxiliary orientation regression task as in \cite{vo}. We also compare our result with another supervised method \cite{Zhai} which learns a cross-view transformation matrix and predict the orientation in a sliding window manner.

Without explicit alignment information from the training set, our method achieves better result than regression and \cite{Zhai} as shown in Fig. \ref{fig:OE} (a). The percentage of samples with error in $[-3.5^{\circ},3.5^{\circ}]$ is around $24\%$ of the whole test set. It is worth noticing that most failure predictions of our method have an error around $180^{\circ}$, because the activation map usually focuses on the road which is symmetrical in aerial view. The same pattern is found in the result of \cite{Zhai}, while the result of regression has a very different pattern. Fewer samples of regression have an error around $180^{\circ}$, while accurate predictions (the central bar on $0^{\circ}$) are also much fewer than the other two methods.
%A possible explanation is that the Euclidean regression loss does not provide enough penalty when the angle error is around $30^{\circ}$.

%To show how different models may affect the orientation estimation performance, 
We further provide the comparison between our baseline and overall framework for orientation estimation in Fig. \ref{fig:OE} (b). Our overall framework leverages superior metric learning techniques, thus achieving better performance than our simple baseline.
%(Section \ref{sec:ablation}).
For example, the percentages of samples with error in $[-3.5^{\circ},3.5^{\circ}]$ are $24\%$ and $21\%$ for our overall and baseline models, respectively. The result shows that better retrieval model does improve the orientation estimation performance.

\subsection{Ablation Study}
\label{sec:ablation}

\iffalse
\begin{figure}[htbp]
\begin{center}
\includegraphics[width=0.8\linewidth]{Figure7_CVUSA-topn.pdf}
\vspace{-0.4cm}
\end{center}
\caption{Top-1 to Top-20 recall comparison of our methods and CVM-Net on CVUSA. }
\label{fig:7}
\end{figure}
\fi

\textbf{Effect of global mining.} To demonstrate the effectiveness of the proposed global mining strategy, we conduct experiments with three settings on our baseline, i.e. no mining, within-batch mining,  and our global mining scheme. As discussed in Section \ref{sec:mining}, the within-batch mining \cite{cvm} or the loss in \cite{reweight} is not able to select hard negative samples, because almost no hard negative samples exists within a mini-batch when the top-$1\%$ accuracy is higher than $80\%$. We report the result of our baseline with within-batch mining (``Batch mining'' in Table \ref{table:mining}) to show the superiority of global mining.  As shown in Table \ref{table:mining}, the global mining significantly improves the performance on both datasets, while the within-batch mining makes little difference on performance. In Fig. \ref{fig:convergence}, we also present the convergence speeds of our baseline and baseline with global mining on CVUSA. The ``Baseline + global mining'' converges much faster than the ``Baseline'', especially in terms of the top-$1$ accuracy. Since our mining strategy aims to find the hardest $r$ samples ($r=5$ in the experiment), it brings relative larger gain on top-$1$ accuracy than top-$1\%$ accuracy. %Given the significant performance improvement, the computational cost of our online update strategy is much less than that of the offline update strategy \cite{netvlad}. %The improvement is also consistent for various top-$n$ measures, as evident from Fig.~\ref{fig:7}.

\begin{table}[htbp]
\begin{center}
\small
\begin{tabular}{l|c|c|c|c}
\hline
\multirow{2}*{Mining Strategy} & \multicolumn{2}{c|}{CVUSA} & \multicolumn{2}{c}{Vo}\\
\cline{2-5}
~ & Top-1\% & Top-1 & Top-1\% & Top-1 \\
\hline
No mining & 96.9\% & 43.7\% & 84.3\% & 7.9\%\\
Batch mining& 96.7\% & 43.0\% & 84.6\% & 8.0\%\\
\hline
Global mining& \textbf{97.0\%} & \textbf{52.1\%}& \textbf{85.8\%} & \textbf{11.1\%}\\
\hline
\end{tabular}
\end{center}
\vspace{-0.5cm}
\caption{Comparison between different mining strategies with our baseline on CVUSA and Vo.}
\label{table:mining}
\end{table}

%-----------------------------------------------------------------------------------------------------
\iffalse
\begin{figure}[htbp]
\vspace{-0.5cm}
\centering
\begin{minipage}[l]{0.48\linewidth}%
%\vspace{1.cm}
    \includegraphics[width=0.99\linewidth]{fig_new/Figure8_convergence.pdf}
\vspace{-0.45cm}
\caption{Top-1 and top-1\% recall accuracy vs. epochs of our baseline w/o and w/ global mining on CVUSA.}
\label{fig:convergence}
\end{minipage}
\hspace{0.1em}
\begin{minipage}[r]{0.48\linewidth}%[htbp]
%\vspace{0.7cm}
\includegraphics[width=.96\linewidth]{fig_new/Figure9_similarity.pdf}
\vspace{-0.3cm}
\caption{Cosine similarity distributions with loss in CVMNet \cite{cvm} (Eq. \ref{eq:2}) and binomial loss (Eq. \ref{eq:4}) on CVUSA.}
\label{fig:distribution}
\end{minipage}
\vspace{-0.2cm}
\end{figure}
\fi
%-----------------------------------------------------------------------------------------------------

\begin{figure}[htbp]
\vspace{-0.5cm}
\centering
    \includegraphics[width=0.75\linewidth]{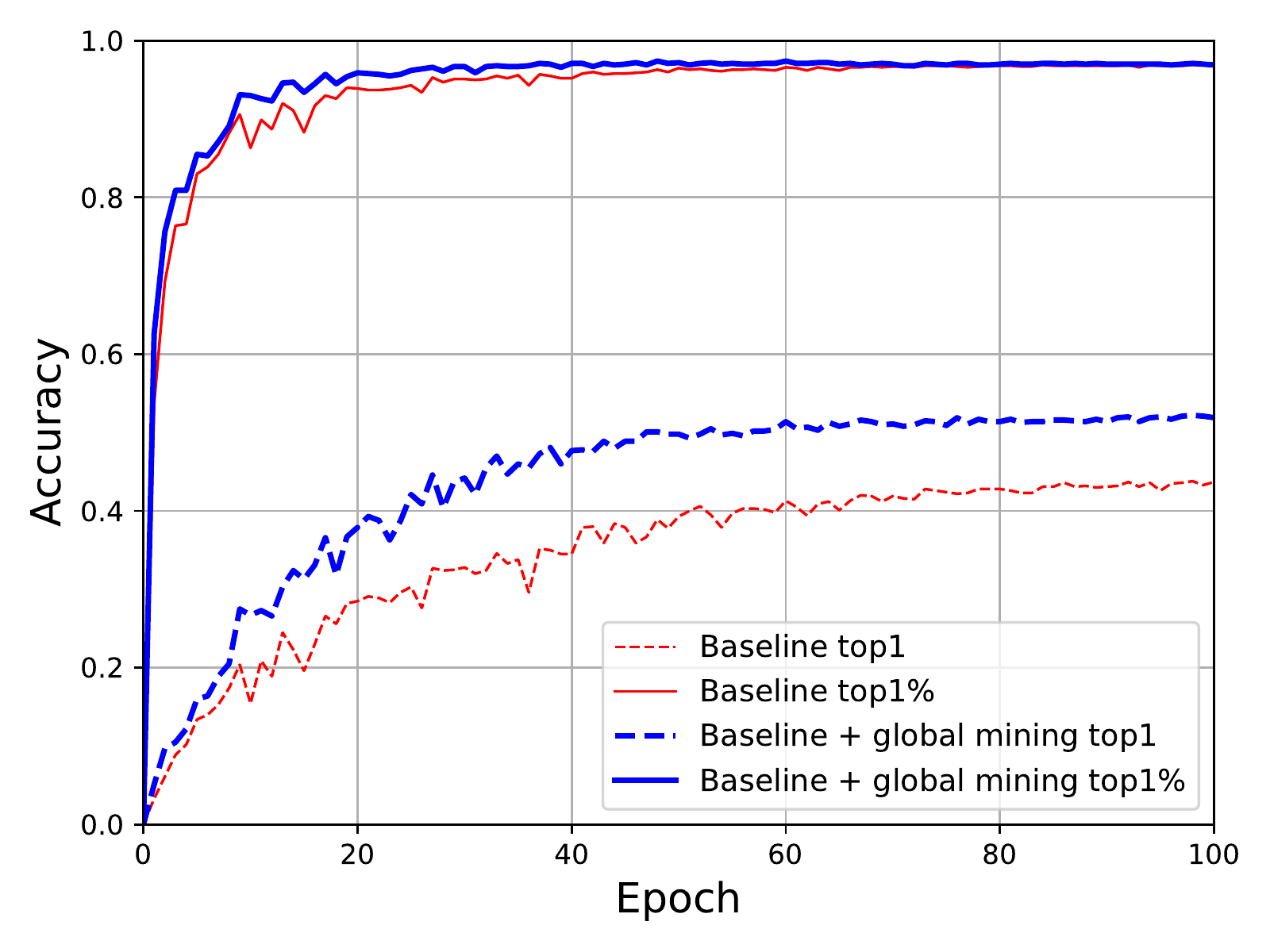}
\vspace{-0.3cm}
\caption{Top-1 and top-1\% recall accuracy vs. epochs of our baseline w/o and w/ global mining on CVUSA.}
\label{fig:convergence}
\end{figure}
\begin{figure}[htbp]
\vspace{-0.5cm}
\centering
\includegraphics[width=0.75\linewidth]{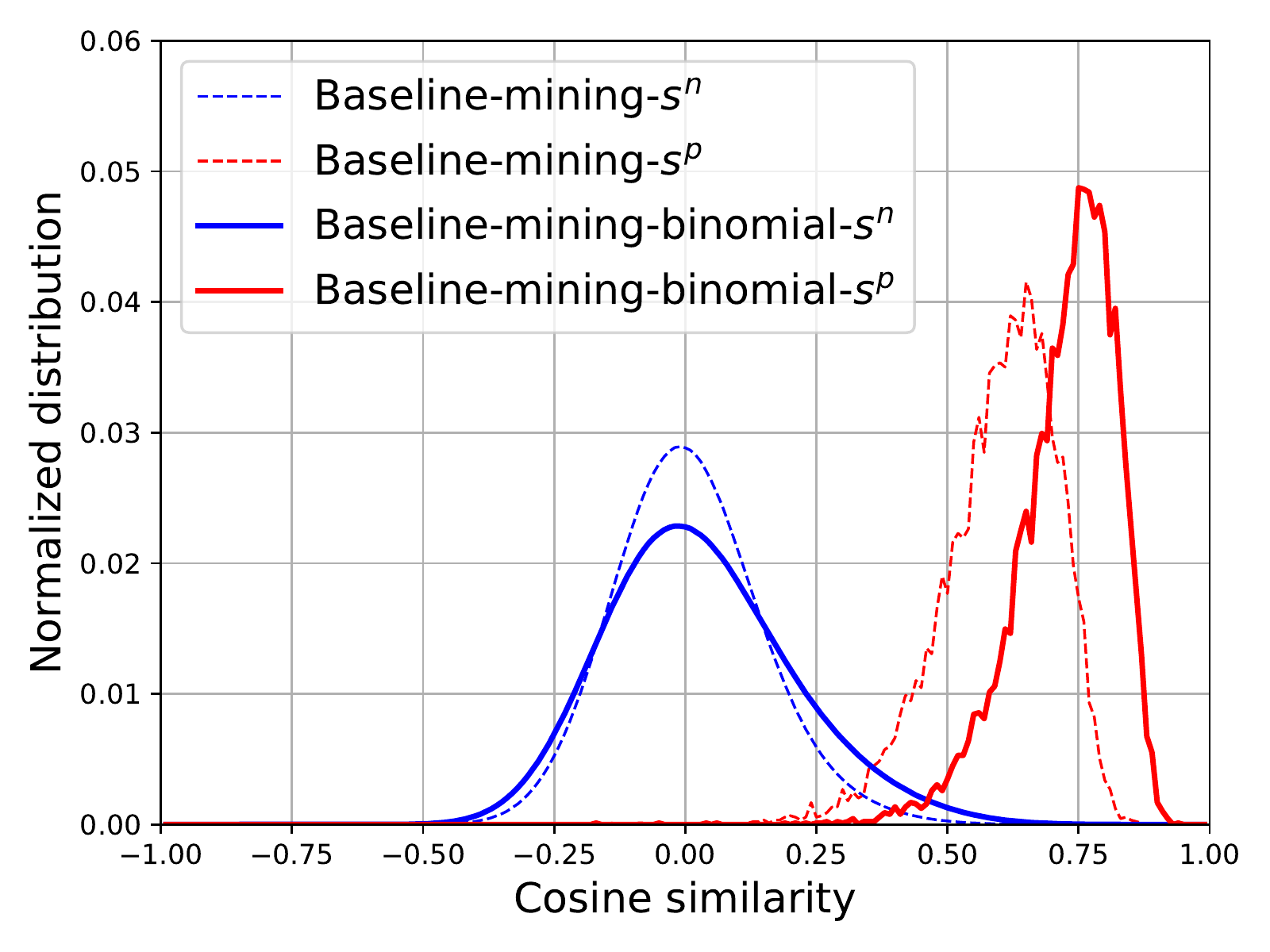}
\vspace{-0.3cm}
\caption{Cosine similarity distributions with loss in CVMNet \cite{cvm} (Eq. \ref{eq:2}) and binomial loss (Eq. \ref{eq:4}) on CVUSA.}
\label{fig:distribution}
%\vspace{-0.2cm}
\end{figure}

\textbf{Effect of binomial loss.} As shown in Fig. \ref{fig:distribution}, the average values of $s^{n}$ and $s^{p}$ (similarity between negative pairs and positive pairs) lie at around $0$ and $0.7$ for ``Baseline + global mining'' with the loss in Eq. \ref{eq:2} \cite{cvm}. An intuitive idea to further improve the performance is breaking the restraint of triplet-like loss which assigns the same weight for positive and negative pairs. As explained in Section \ref{sec:loss}, the binomial loss (Eq. \ref{eq:4}) puts a stronger constraint on positive samples (larger mean value and smaller variance in Fig. \ref{fig:distribution}), while the distribution of negative pairs is more scattered (larger variance). This characteristic is beneficial to the performance on both datasets as shown in Table \ref{table:loss}. %Different from the effect of global mining which has a large impact on top-$1$ accuracy, the change of distribution benefits more on top-$1\%$ accuracy. The top-$1\%$ accuracy on CVUSA is improved by $0.7\%$ even when the accuracy is over $97\%$. %We also experiment with loss in Eq. \ref{eq:4} adopting the same weight ($\alpha_n = \alpha_p$) (``Binomial-Same'' in Table \ref{table:loss}). ``Binomial'' with different weights outperforms ``Binomial-Same'' which supports our analysis about the sample imbalance between positive and negative samples.
\begin{table}[htbp]
\small
%\vspace{-0.4cm}
\begin{center}
\begin{tabular}{l|c|c|c|c}
\hline
\multirow{2}*{Loss Function} & \multicolumn{2}{c|}{CVUSA} & \multicolumn{2}{c}{Vo}\\
\cline{2-5}
~ & Top-1\% & Top-1 & Top-1\% & Top-1 \\
\hline
CVMNet \cite{cvm} & 97.0\% & 52.1\% & 85.8\% & 11.1\% \\
%Binomial-Same & 98.5\% & 62.9\% & 86.8\% & 10.3\% \\
\hline
Binomial (Eq. \ref{eq:4}) & \textbf{97.7\%} & \textbf{54.5\%} & \textbf{88.3\%} &\textbf{11.8\%}\\
\hline
\end{tabular}
\end{center}
\vspace{-0.3cm}
\caption{Comparison between different loss functions with our baseline and global mining on CVUSA and Vo.}
\vspace{-0.4cm}
\label{table:loss}
\end{table}

\iffalse
\begin{figure}[htbp]
\begin{center}
\includegraphics[width=0.8\linewidth]{fig_new/Figure10_Vo-topn.pdf}
\vspace{-0.4cm}
\end{center}
\caption{Top-$n$ recall accuracy of our methods w/o overlapping training set on Vo and Hays.}
\label{fig:10}
\end{figure}
\fi

\textbf{Effect of alignment.} As discussed in Section \ref{sec:alignment}, the alignment setting has a large impact on the performance, we thus report the ablation study results of CVUSA on both settings (w/ or w/o alignment) in Table \ref{table:alignment}. The ``Overall'' denotes our baseline with global mining and binomial loss in Eq. \ref{eq:4}. As expected, the improvements of the proposed techniques are consistent across both settings. Apparently, training with alignment can improve the retrieval performance, because it assumes the inference images to be aligned, but this may be infeasible for challenging real-world applications.
%Since our framework is independent on the alignment of training set, our baseline may serve as a general cross-view image matching pipeline. 
%We recommend future works to report the results on both settings. 

%\vspace{1.8cm}
\begin{table}[htbp]
%\vspace{-0.5cm}
\footnotesize
\begin{center}
\begin{tabular}{l|c|c|c|c}
%\begin{tabular}{p{1.85cm}|p{1.0cm}|p{1.0cm}|p{1.0cm}|p{1.0cm}}
\hline
\multirow{2}*{Method} & \multicolumn{2}{c|}{w/ alignment} & \multicolumn{2}{c}{w/o alignment}\\
\cline{2-5}
~ & Top-1\% & Top-1 & Top-1\% & Top-1 \\
\hline
Baseline & 98.8\% & 60.1\% & 96.9\% & 43.7\% \\
Baseline+global mining & 98.8\% & 67.0\% & 97.0\% & 52.1\% \\
Overall & \textbf{99.1\%} & \textbf{70.4\%} & \textbf{97.7\%} &\textbf{54.5\%}\\
\hline
\end{tabular}
\end{center}
\vspace{-0.3cm}
\caption{\small{Comparison between our methods w/ and w/o alignment (random rotated aerial images) on CVUSA. ``Overall" = Baseline+global mining+binomial loss (Eq. \ref{eq:4})}.}
\vspace{-0.4cm}
\label{table:alignment}
\end{table}

\section{Conclusion}
In this paper, we revisit cross-view image geo-localization and orientation estimation and highlight the effect of image alignment information which is usually ignored by previous works.
%We clearly illustrate the effect of alignment on performance and beyond with ablation study and visual explanation. 
Our analysis indicates the alignment has a great impact on the retrieval performance.
%comparison without the claim of alignment setting can be unfair for previous works. 
Furthermore, we identify the unique challenges of geo-localization and propose a global mining strategy along with the binomial loss to tackle them. Extensive experiments on two widely used benchmark datasets show the superiority of the proposed method on both alignment settings (w or w/o). Moreover, our model trained without alignment is able to predict the orientation (angle) of a cross-view image pair without any alignment information supervision during the training. The proposed orientation estimation method achieves state-of-the-art result on CVUSA.

{\small
\bibliographystyle{ieee_fullname}
\bibliography{egbib}
}

\clearpage
\appendix
\section{Appendix}
In the appendix, we provide more results and analysis of our method including:
\begin{itemize}
    \item Top-$n$ accuracy distributions on the CVUSA and Vo datasets.
    \item Retrieval case study on both datasets.
    \item Additional qualitative results of Grad-CAM on both datasets.
    \item Orientation estimation case study on CVUSA dataset.
\end{itemize}

\vspace{0.1cm}

\subsection{Top-$n$ Accuracy}
\begin{figure}[hb]
\begin{center}
\includegraphics[width=0.98\linewidth]{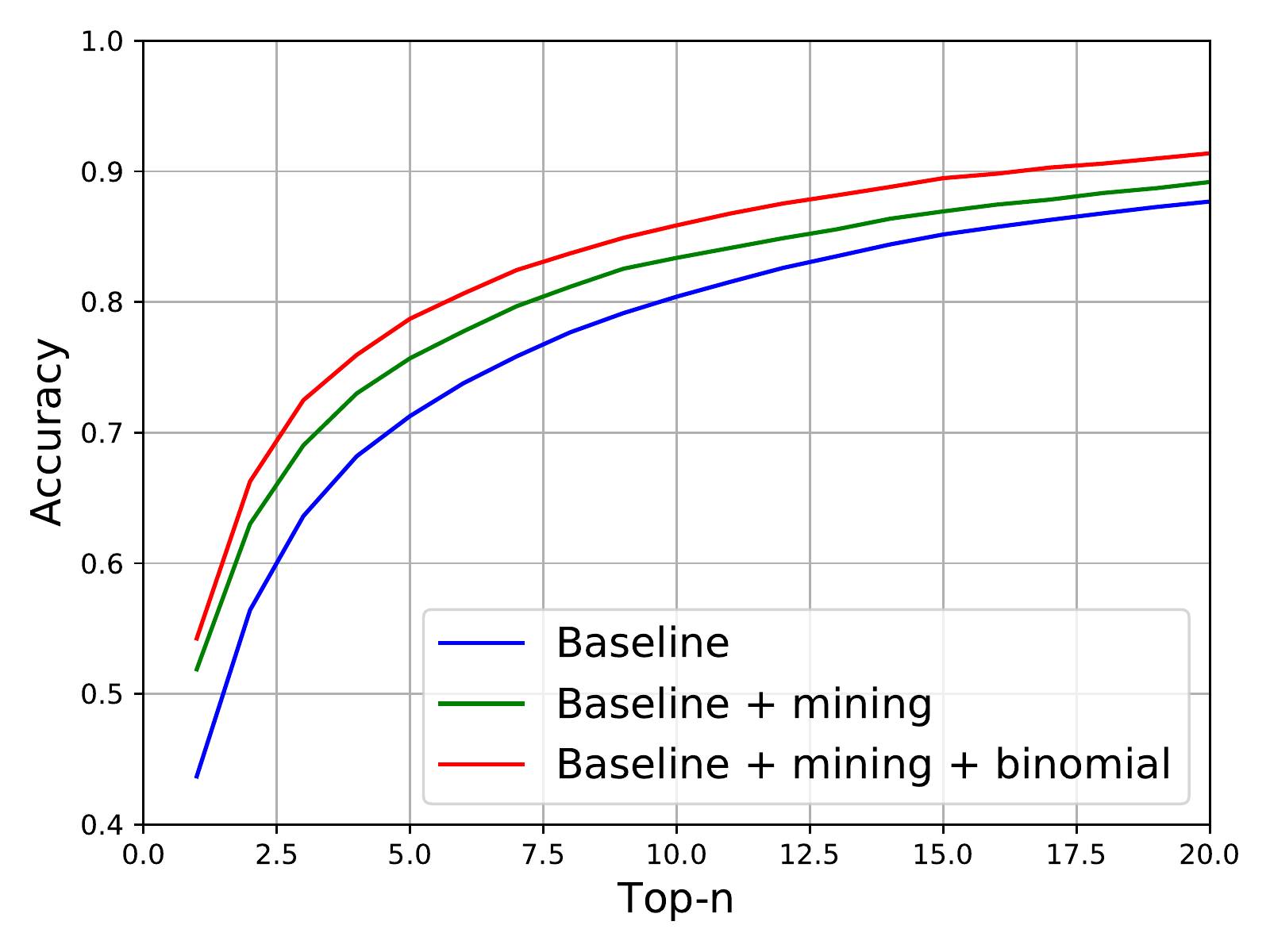}
\end{center}
\vspace{-0.5cm}
\caption{Top-$n$ accuracy on CVUSA dataset.}
\label{fig:top-n-cvusa}
\end{figure}

\paragraph{\bf CVUSA}
Fig. \ref{fig:top-n-cvusa} shows the top-$1$ to top-$20$ accuracy of our methods on CVUSA without image alignment, i.e. randomly rotated aerial images in the training phase. These methods include Baseline (``Baseline''), Baseline + global mining (``Baseline + mining''), and Baseline with global mining and binomial loss (``Baseline + mining + binomial'').

\begin{figure}[htbp]
\begin{center}
\includegraphics[width=0.98\linewidth]{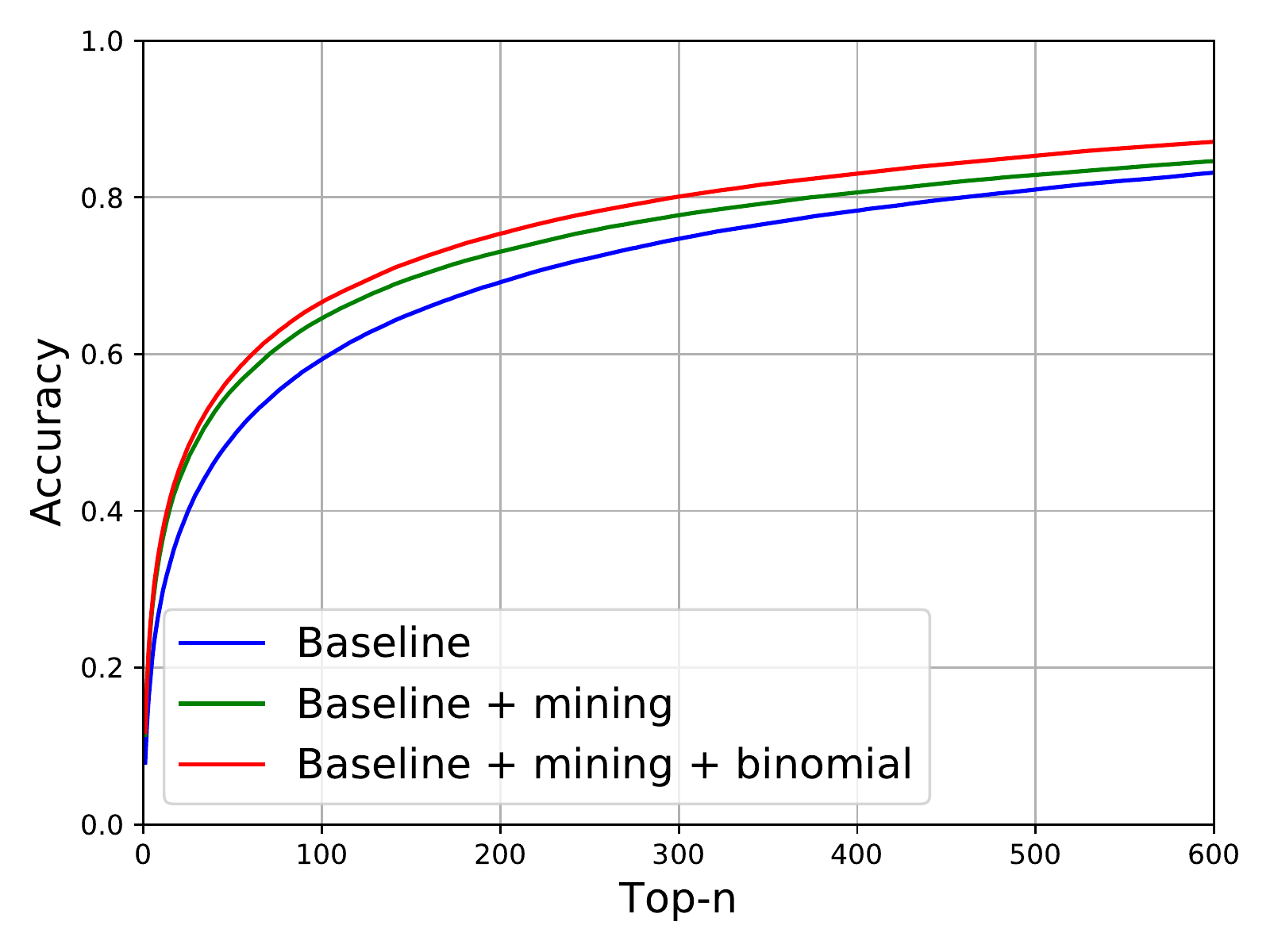}
\end{center}
\vspace{-0.5cm}
\caption{Top-$n$ accuracy on Vo dataset.}
\label{fig:top-n-vo}
\end{figure}

\paragraph{\bf Vo}
Fig. \ref{fig:top-n-vo} shows the top-$1$ to top-$600$ accuracy of our methods on Vo dataset without image alignment, i.e. Baseline (``Baseline''), Baseline + global mining (``Baseline + mining''), and Baseline with global mining and binomial loss (``Baseline + mining + binomial'').
\vspace{0.2cm}

\subsection{Retrieval Results}
\paragraph{{\bf CVUSA}} To further illustrate our matching results, we present a few retrieval examples in Fig.~\ref{fig:case}. Specifically, in this figure,  the first column lists 4 street-view query images from CVUSA. The second column shows the corresponding matching aerial images (ground truth). Their rankings (1, 5, 99 and 957) in our retrieval outcome are provided on top of each image. The third to the last columns are the top-5 retrieved aerial images for each query.

\vspace{0.4cm}
The similarity scores (``probability'' in Fig. \ref{fig:case}) are also provided for all retrieved images. The successful case ranked at 1 (the first query example in the first row) contains many objects, e.g. houses with distinguishable features.
However, the second query image (second row) only contains road and trees which are similar in all images. As shown in its retrieved aerial images, all the top 4 samples/images can be considered as hard negatives for this anchor. Poor cases in the third and fourth rows also contain no distinguishable objects. Other possible explanation for the poor retrieval results is the dramatic appearance difference between two views (may be acquired at different seasons or illumination conditions).
\begin{figure*}[htbp]
\centering
\includegraphics[width=0.99\linewidth]{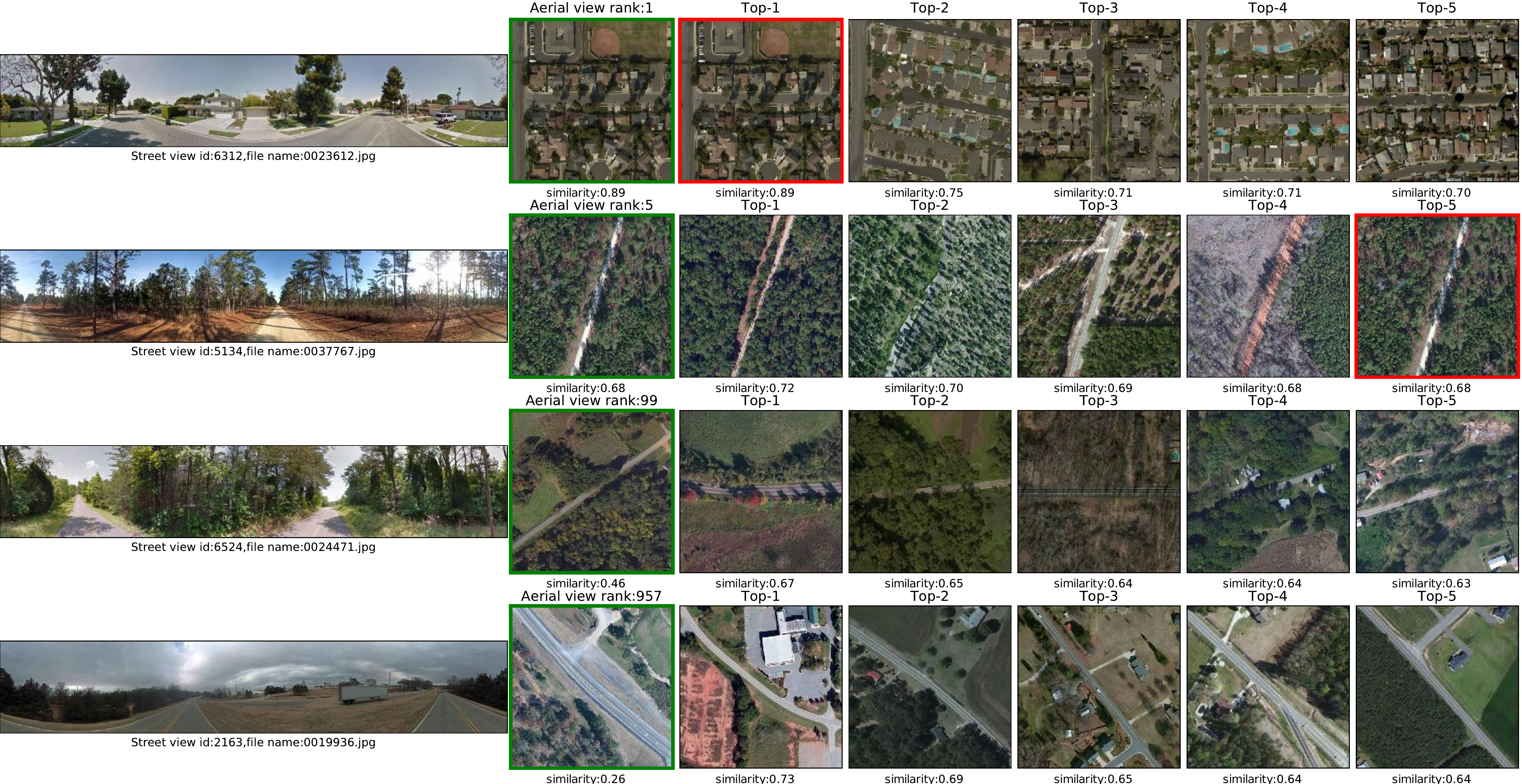}
\caption{Image retrieval examples of our method on CVUSA dataset. For each query street-view image (first column), we show the top-5 retrieved aerial images (third to the last columns). The ground truth is marked in green box (i.e. second column). The red box marks the same aerial image as the ground truth. The last two rows show two examples where the ground truth images are not in the top-5 (i.e. rank $=99$ and rank $=957$) of the retrieved results.}
\label{fig:case}
\end{figure*}

\paragraph{\bf Vo} Similarly, Fig.~\ref{fig:1} shows four cases of our method on Vo dataset with ranking and similarity of the paired positive aerial image/sample as well as the retrieved top-5 samples. The first row shows a successful geo-localization case with its positive sample ranking at top-1. The other three rows show the cases in which the positive sample is ranked at 5, 600 (about top 1\%), and 5984 (about top 10\%), respectively. 
\begin{figure*}[htbp]
\begin{center}
\includegraphics[width=0.99\linewidth]{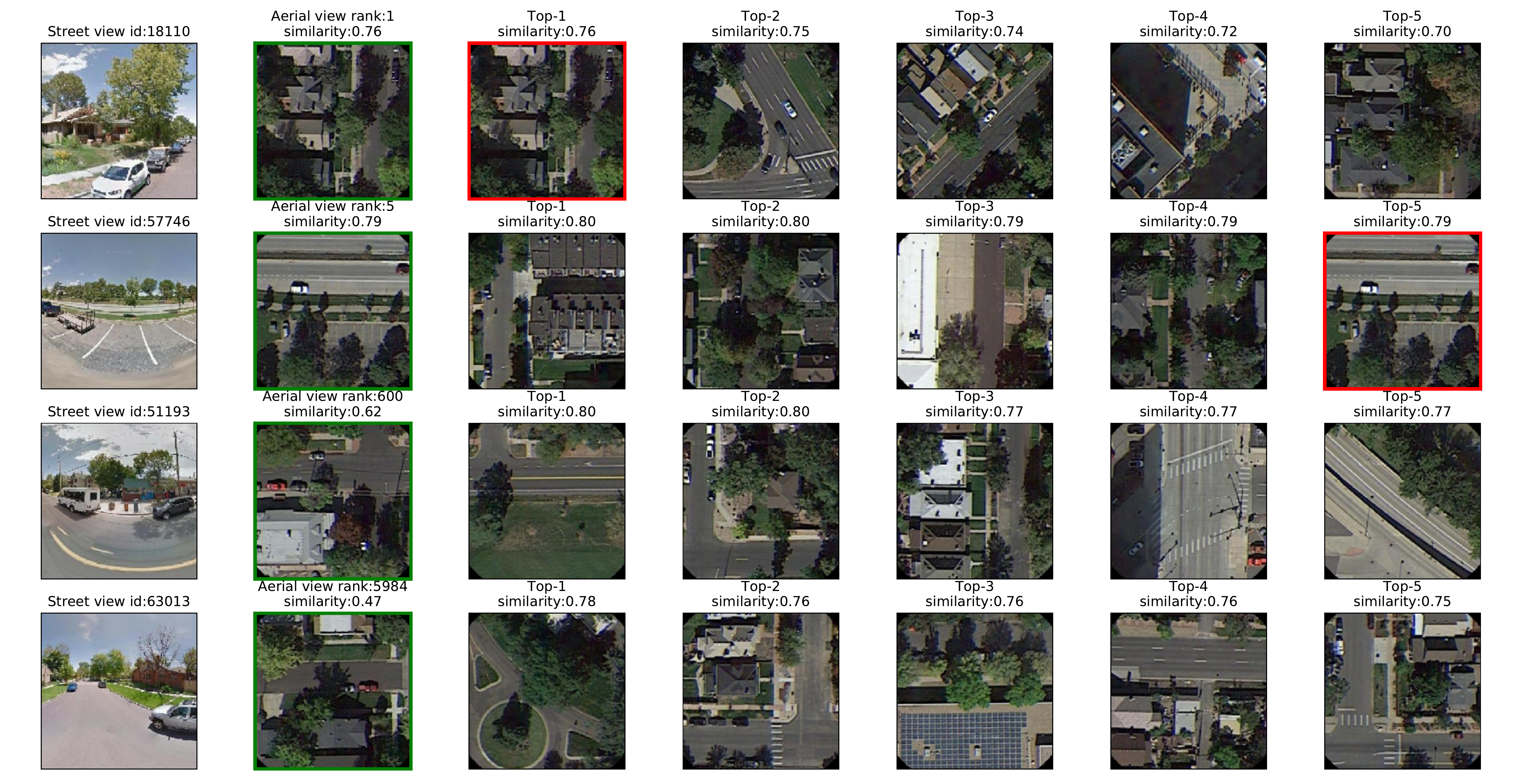}
\end{center}
\vspace{-0.1cm}
\caption{Example retrieval results of our method on Vo  dataset. The ground truth is marked in green box (i.e. second column). The red box marks the same aerial image as the ground truth.}
\label{fig:1}
\end{figure*}
% %

\clearpage
\onecolumn
\subsection{Grad-CAM Visualization Result for Cross-View Image Matching}
\begin{figure*}[h]
\begin{center}
\includegraphics[width=0.85\linewidth]{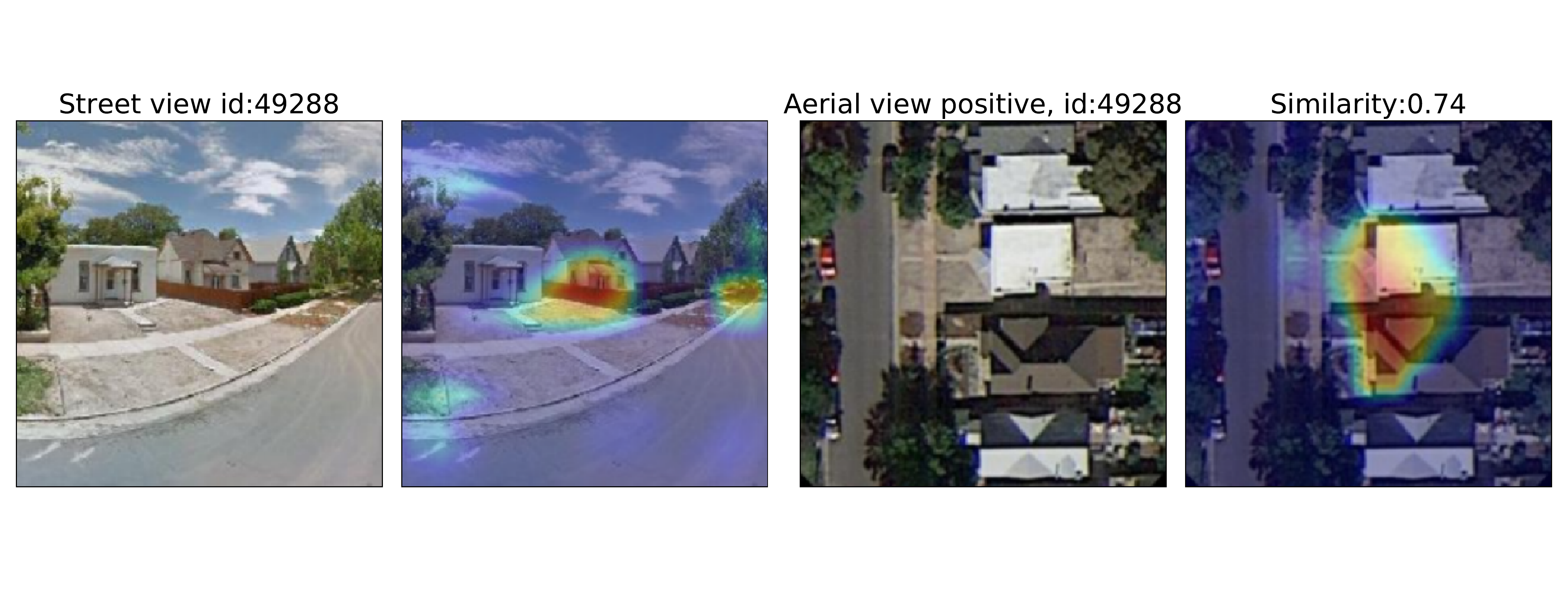}\vspace{-1cm}
\includegraphics[width=0.85\linewidth]{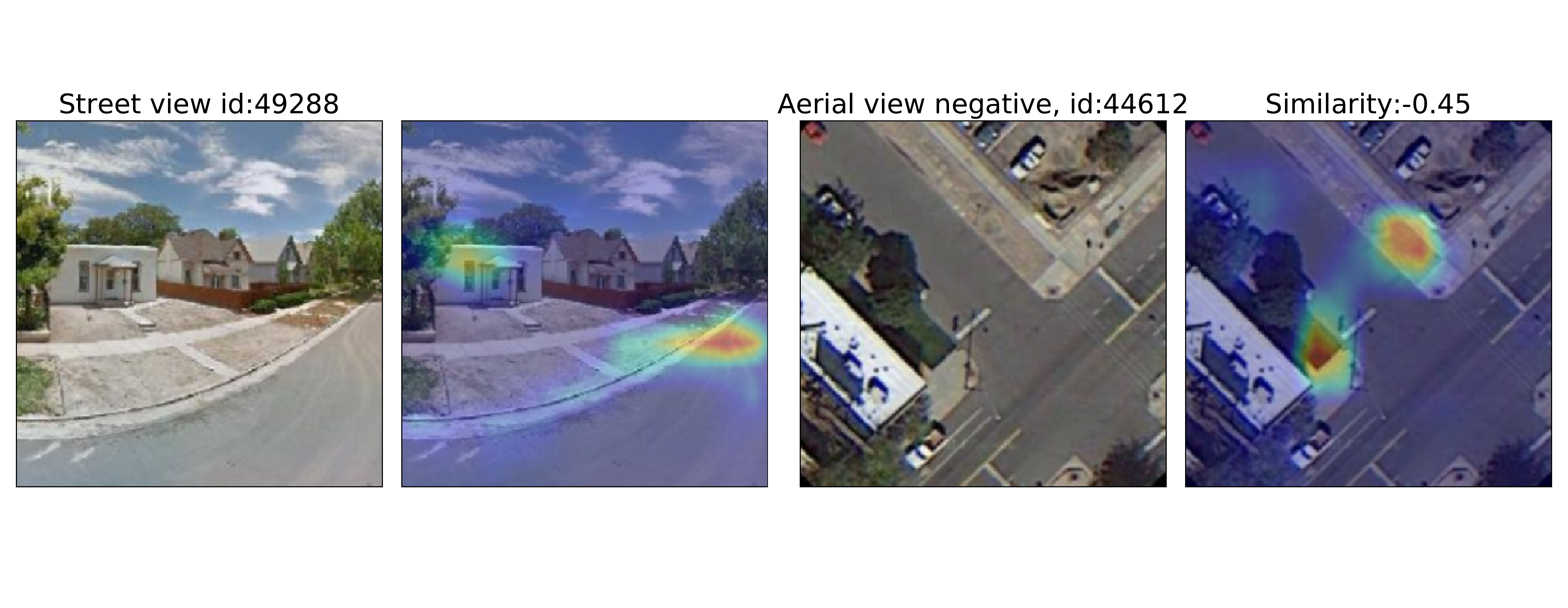}
\end{center}
\vspace{-1.5cm}
\caption{Grad-CAM generated from positive and negative image pairs of the Vo and Hays dataset.}
\label{fig:3}
\end{figure*}

Figs.~\ref{fig:2} and \ref{fig:3} show the examples of activation map generated by Grad-CAM on CVUSA and Vo datasets, respectively. Each figure contains the original images from two views and the activation maps when the street-view image is matched with a positive sample and a negative sample. \textbf{The same street-view image has different activation maps when matched with different aerial images}. For example, in Fig.~\ref{fig:2}, the activation map of the query street-view image focuses on the buildings when matched with the paired/matching aerial image (positive sample) which also contains many buildings/houses, while it focuses on the road when matched with a negative sample containing no building/house.
\begin{figure*}[htbp]
\vspace{-0.3cm}
\begin{center}
\includegraphics[width=0.85\linewidth]{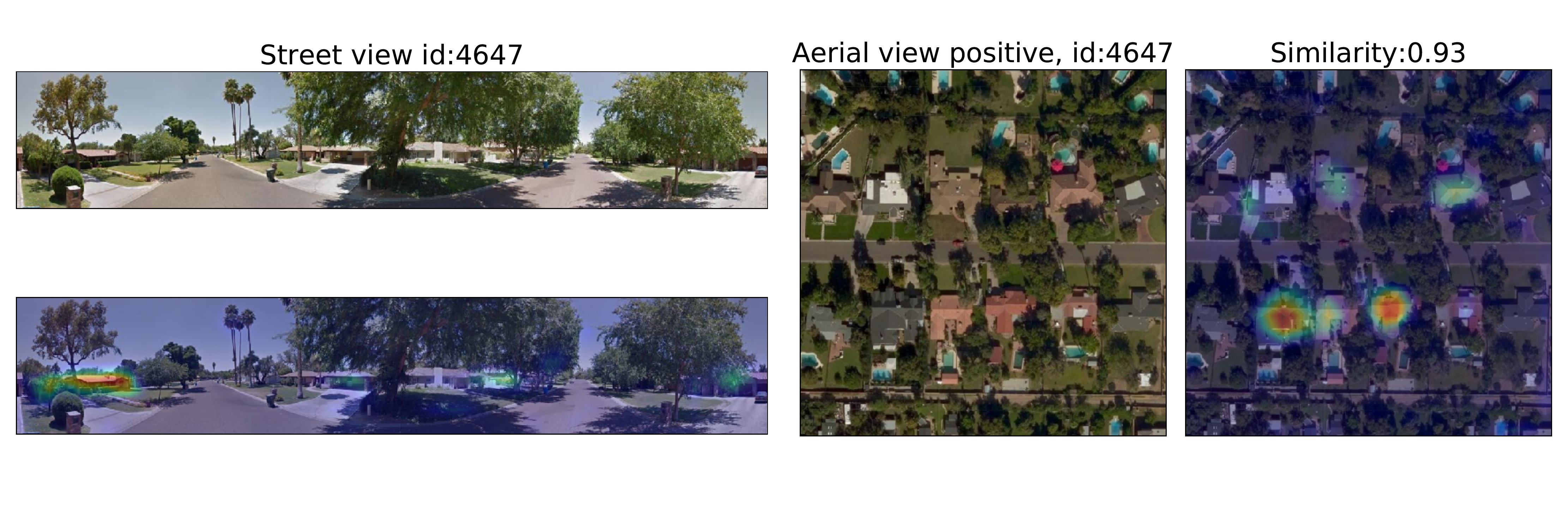}\vspace{-1cm}
\includegraphics[width=0.85\linewidth]{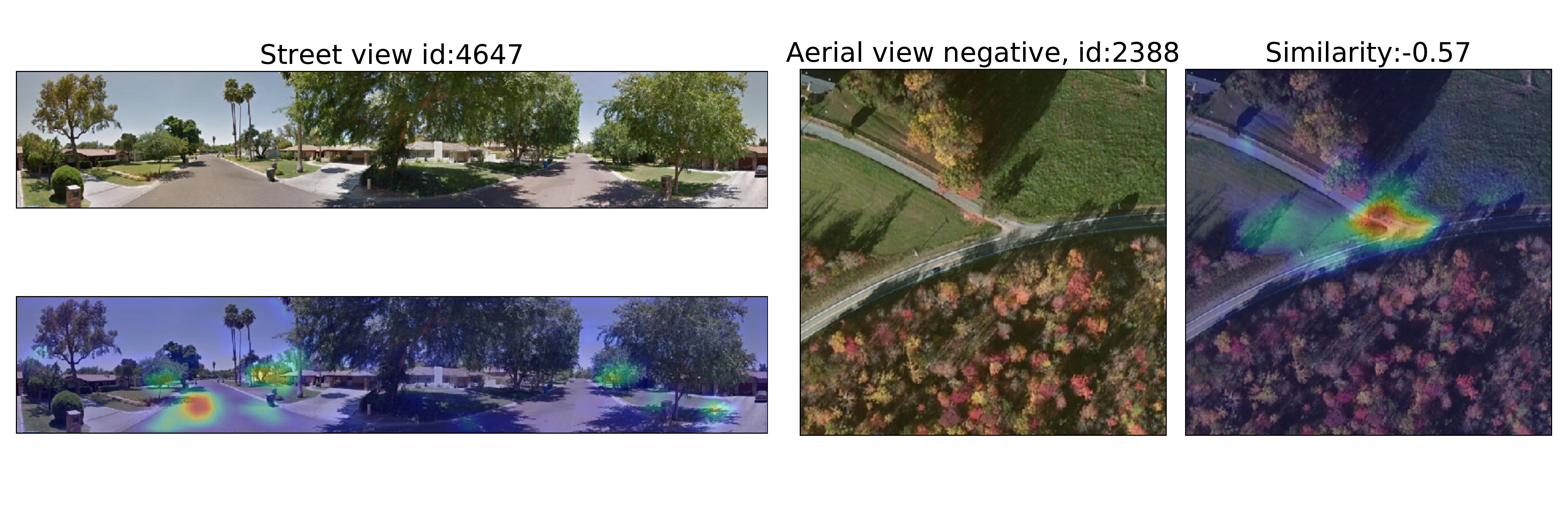}
\end{center}
\vspace{-1cm}
\caption{Grad-CAM generated from positive and negative image pairs of the CVUSA dataset.}
\label{fig:2}
\end{figure*}
\vspace{-0.9cm}

\clearpage
\subsection{Orientation Estimation Examples}
\begin{figure*}[h]
\begin{center}
\includegraphics[width=0.99\linewidth]{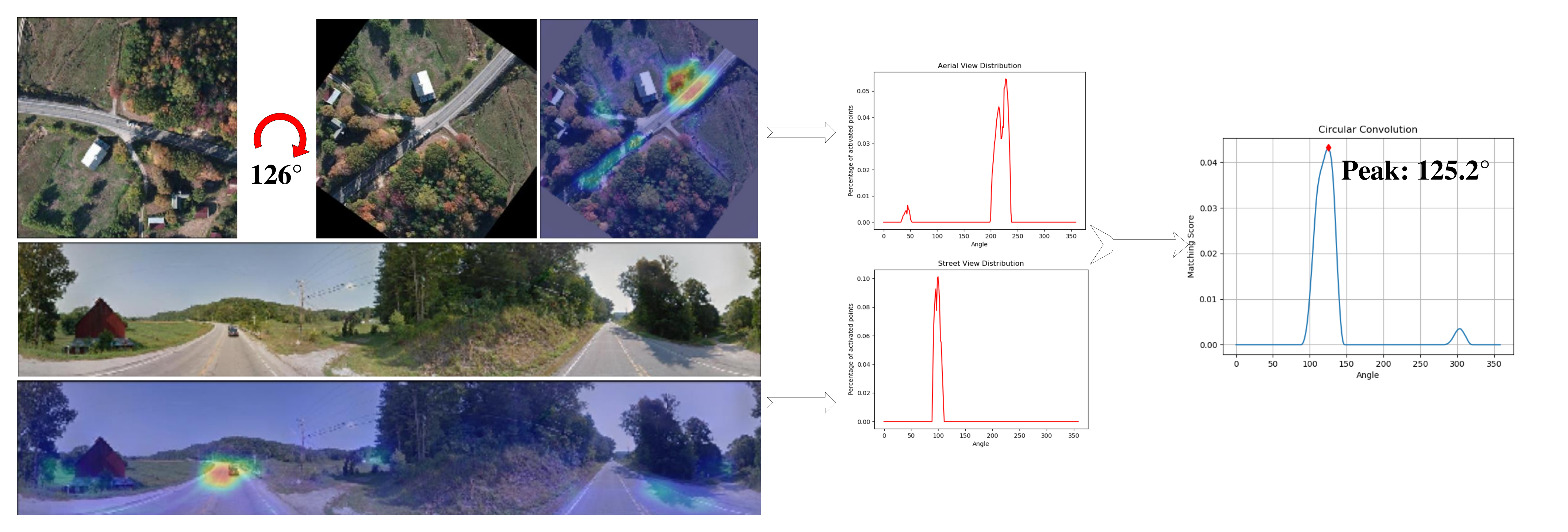}
\vspace{-0.3cm}
\end{center}
\caption{Successful orientation estimation between a pair of cross-view images.}
\label{fig:5_good}
\end{figure*}

\begin{figure*}[h]
\begin{center}
\includegraphics[width=0.99\linewidth]{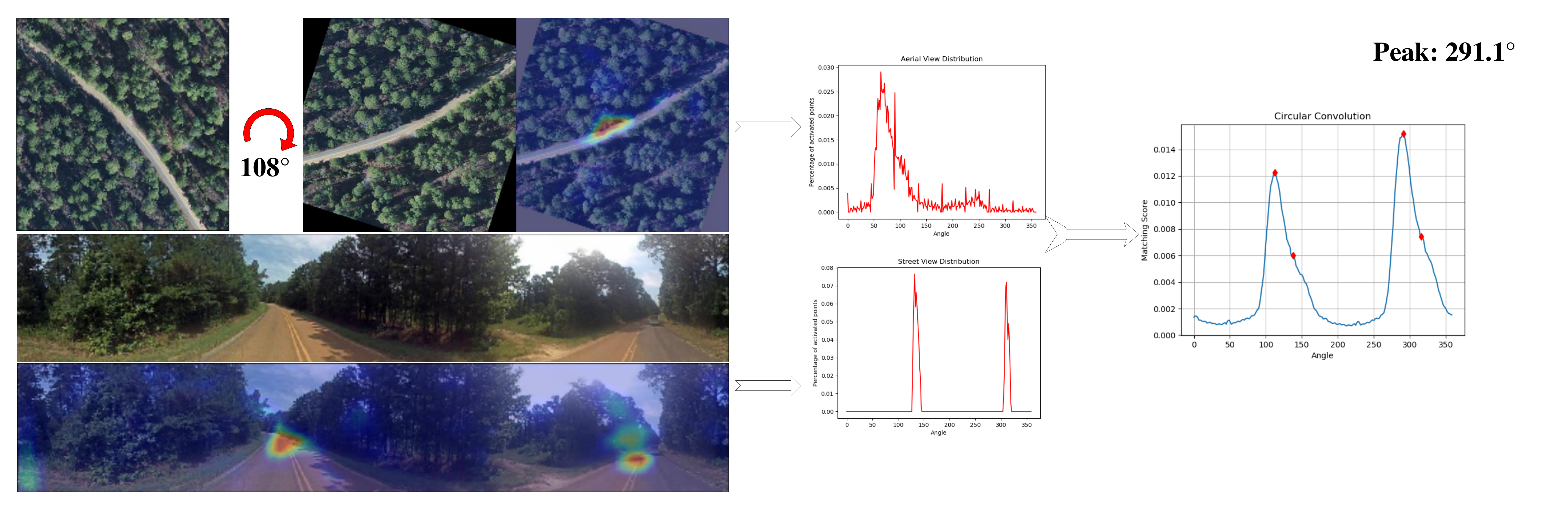}
\vspace{-0.3cm}
\end{center}
\caption{Failure orientation estimation between a pair of cross-view images.}
\label{fig:5_bad}
\end{figure*}

Figs.~\ref{fig:5_good} and \ref{fig:5_bad} respectively show a successful and a failure prediction of orientation (i.e. angle) between a pair of cross-view images using our proposed method, i.e. matching the randomly rotated aerial image with panorama street-view image and using the Grad-CAM activation map for orientation estimation. In Fig.~\ref{fig:5_bad}, due to the symmetry of the road, the result of circular convolution is a distribution with multiple peaks. The highest peak is not the best one and the angle error is about $180^{\circ}$. In fact, the second peak is a better prediction for this example.

%\vspace{-1cm}

\end{document}